\documentclass[]{jingdong}
\usepackage[export]{adjustbox}
\usepackage[toc,page,header]
{appendix}
\usepackage{minitoc}
\usepackage{url}
\usepackage{amssymb}
\usepackage[utf8]{inputenc}
\usepackage{microtype}
\usepackage{booktabs}
\usepackage{pifont} 
\usepackage{multirow}
\usepackage{makecell}
\usepackage{paralist}
\usepackage{xspace}
\usepackage{color}
\usepackage{xcolor}
\usepackage{colortbl}
\usepackage{adjustbox}
\usepackage{hyperref} 
\usepackage[edges]{forest}
\usepackage{tikz} 
\usepackage{caption}
\usepackage{amsfonts}
\usepackage[toc,page,header]{appendix}
\usepackage[utf8]{inputenc}
\usepackage{bbm}
\usepackage{enumitem}
\usepackage{pgfplots}
\pgfplotsset{compat=1.18}

\usepackage{subcaption}   
\usepackage{pifont}

\usepackage{threeparttable}  
\usepackage{graphicx}  

\definecolor{seedc}{RGB}{7, 92, 173}

\newcommand{\name}[1]{GM-0}

\newcommand{\hardware}[1]{ByteMini}

\renewcommand{\paragraph}[1]{\vspace{0.1em}\noindent\textbf{#1}}

\usepackage{tabularx}       
\usepackage{latexsym}
\usepackage[T1]{fontenc}
\usepackage[utf8]{inputenc}
\usepackage{microtype}
\usepackage{inconsolata}
\usepackage{graphicx}
\usepackage{hyperref}       
\usepackage{url}            
\usepackage{booktabs}       
\usepackage{amsfonts}       
\usepackage{nicefrac}       
\usepackage{stackengine}
\usepackage{microtype}      
\usepackage{colortbl}
\usepackage{xcolor}
\usepackage{amsmath}
\usepackage{amssymb}
\usepackage{amsthm}
\usepackage{mathrsfs}
\usepackage{pifont}
\usepackage{MnSymbol}
\usepackage{balance}
\usepackage{enumitem}
\usepackage{listings}
\usepackage{xcolor}
\usepackage{natbib}
\usepackage{multicol}

\AtBeginDocument{%
  \providecommand\BibTeX{{%
    \normalfont B\kern-0.5em{\scshape i\kern-0.25em b}\kern-0.8em\TeX}}}

\makeatletter
\DeclareRobustCommand\onedot{\futurelet\@let@token\@onedot}
\def\@onedot{\ifx\@let@token.\else.\null\fi}

\usepackage{setspace}
\usepackage{mathtools}

\usepackage{multirow,booktabs}
\usepackage{subcaption}

\newcommand{\owo}[1]{\textsc{OAgents}}

\definecolor{lightgreen}{RGB}{144, 238, 144} 
\definecolor{lightred}{RGB}{255, 105, 97}

\newtcolorbox{promptbox}[2][Prompt]{
colback=black!5!white,
arc=5pt, 
boxrule=0.5pt,
fonttitle=\bfseries,
title=#1, 
before upper={\small}, fontupper=\fontfamily{ptm}\selectfont,
colframe=#2, 
}
\definecolor{ogreen}{RGB}{34, 139, 34} 

\usepackage{cleveref}
\theoremstyle{plain}

\theoremstyle{definition}

\theoremstyle{remark}
\usepackage{booktabs}
\usepackage{longtable}
\usepackage{svg}
\usepackage{subcaption}
\usepackage{tikz}

\usepackage[textsize=tiny]{todonotes}
\usepackage{fontawesome}
\usepackage{wrapfig}

\title{JoyAI-Sim: A Simulation-Enabled Interconversion Toolchain for the Embodied Data Pyramid}
\abstract{
Generalist robot policies require trustworthy evaluation and robot-centered
training data, but both are difficult to scale with physical robots alone.
Real-robot trials and demonstrations remain the most faithful source of
deployment signals,
yet they are often slow,
costly,
and hard to reproduce.
We present \textbf{JoyAI-Sim},
a simulation-enabled interconversion toolchain for human-robot aligned model evaluation and data generation,
denoted as \textbf{Robot $\rightleftharpoons$ Simulation $\rightleftharpoons$ Human}.
On the one hand, the \textbf{Robot $\rightarrow$ Simulation $\rightarrow$ Human} pathway supports human-robot aligned model evaluation by reconstructing real-robot tabletop organization tasks as calibrated digital twins for scalable evaluation, while using human embodied feedback to inspect and refine the naturalness of simulated motions.
On the other hand, the \textbf{Human $\rightarrow$ Simulation $\rightarrow$ Robot} pathway supports human-robot aligned data generation: it lifts
egocentric human demonstrations into simulation,
checks them under robot physical constraints,
and converts them into robot-centered trajectories,
annotations,
and visual observations.
Together,
these pathways use the simulator as both a scalable evaluation layer and a
physical consistency filter for robot data generation.
We further package the core reconstruction, simulation,
rendering,
and realism-augmentation modules as cloud services on JD Cloud,
turning the system into a reusable and scalable infrastructure for robot data generation and model evaluation.
}

\checkdata[Project Page]{\url{https://joyai-sim.github.io/}}

\author[]{\small Peidong~Liu, Yongce~Liu, Songyan~Guo, Fuyuan~Ma, Zhihao~Yuan, Ao~Li, Zengjue~Chen, Wenhao~Li, Tianle~Zhang, Mingyang~Li, Jiale~Zhang, Junzhe~Xiong, Zhiyuan~Xiang, Dafeng~Chi, Yuzheng~Zhuang, Liyi~Luo, Wei~Tan, Dongjiang~Li, Nan~Jiang, Yihang~Li, Qingrong~He, Jiaming~Liang, Chen~Cai, Mingxi~Luo, Hui~Zhang, Peng~Hao, Song~Wang, Ning~Qiao, Yince~Gao, Lei~Kang, Junwu~Xiong, Ruodai~Li, Jiawei~Li\textsuperscript{\dagger}, Hui~Shen, Yicheng~Gong, Nan~Duan, Liang~Lin\textsuperscript{\dagger}}

\affiliation[]{
\large
\begin{tabular}{c}
Joy Future Academy, JD Group \\
JD Technology, JD Group
\end{tabular}
}

\begin{document}
\maketitle

\begingroup
\renewcommand{\thefootnote}{\dagger}
\footnotetext{Corresponding authors: Jiawei Li <li-jw15@tsinghua.org.cn>, Liang Lin <linliang@ieee.org>.}
\endgroup

\section{Introduction}
\label{sec:introduction}

Generalist robot policies are increasingly expected to operate reliably in
complex manipulation settings~\cite{brohan2023rt,kim2024openvla,
zhang2026joyaira01foundationmodel}.
Progress toward this goal depends on two core resources: trustworthy
evaluation and robot-centered training data.
Real-robot trials and demonstrations remain the most faithful source for both,
because they expose the full deployment stack. However, relying primarily on
physical robots creates a severe scaling problem. Trials are slow and hard to
reproduce; demonstrations require specialized hardware and repeated scene
resets; both require safety monitoring to avoid hardware damage and unsafe
interactions. As a result, simulation and human demonstrations are natural
alternatives for scaling evaluation and data collection, but using them
effectively requires bridging a gap between scalability and deployment
faithfulness~\cite{oneill2024openx,khazatsky2024droid,walke2023bridgedatav2}.

This scaling gap creates two complementary bottlenecks. The first is an
evaluation bottleneck. Simulation offers scalable execution,
controllable reset, privileged state, and parallel rollouts, but it cannot be
treated as a black-box substitute for physical deployment~\cite{
james2019rlbench,chen2025robotwin,nasiriany2024robocasa}. Fragile
initialization, inaccurate assets or physics, and success checkers that do
not match task semantics can change measured success rates independently of
policy quality. A practical evaluation pipeline should therefore start from
real-world tasks and success criteria, reconstruct them as calibrated digital
twins, and use simulation as a scalable screening layer before final physical
validation. The second is a data bottleneck. Human egocentric videos
are abundant and diverse, and they cover a large range of everyday
manipulation behaviors, but they are not directly executable by robots because
human hands and robot end-effectors obey different kinematic, contact, and
control constraints~\cite{grauman2022ego4d,
li2026egolive,pavlakos2024hamer,hsieh2025dexman}.
Simulation can serve as the missing middle layer: human motion and task
scenes can be reconstructed, checked under robot physical constraints, and
converted into robot-centered trajectories and observations.
In this role, simulation is not merely an evaluation environment, but also a
physical consistency filter and data amplifier. We refer to this hierarchy as the embodied data pyramid: robot data provides the most deployment-faithful but scarce supervision, simulation data provides scalable and physically inspectable intermediate representations, and human data provides abundant but embodiment-mismatched observations, demonstrations, and task-level priors.

\begin{figure}[!t]
    \centering
    \includegraphics[width=\linewidth]{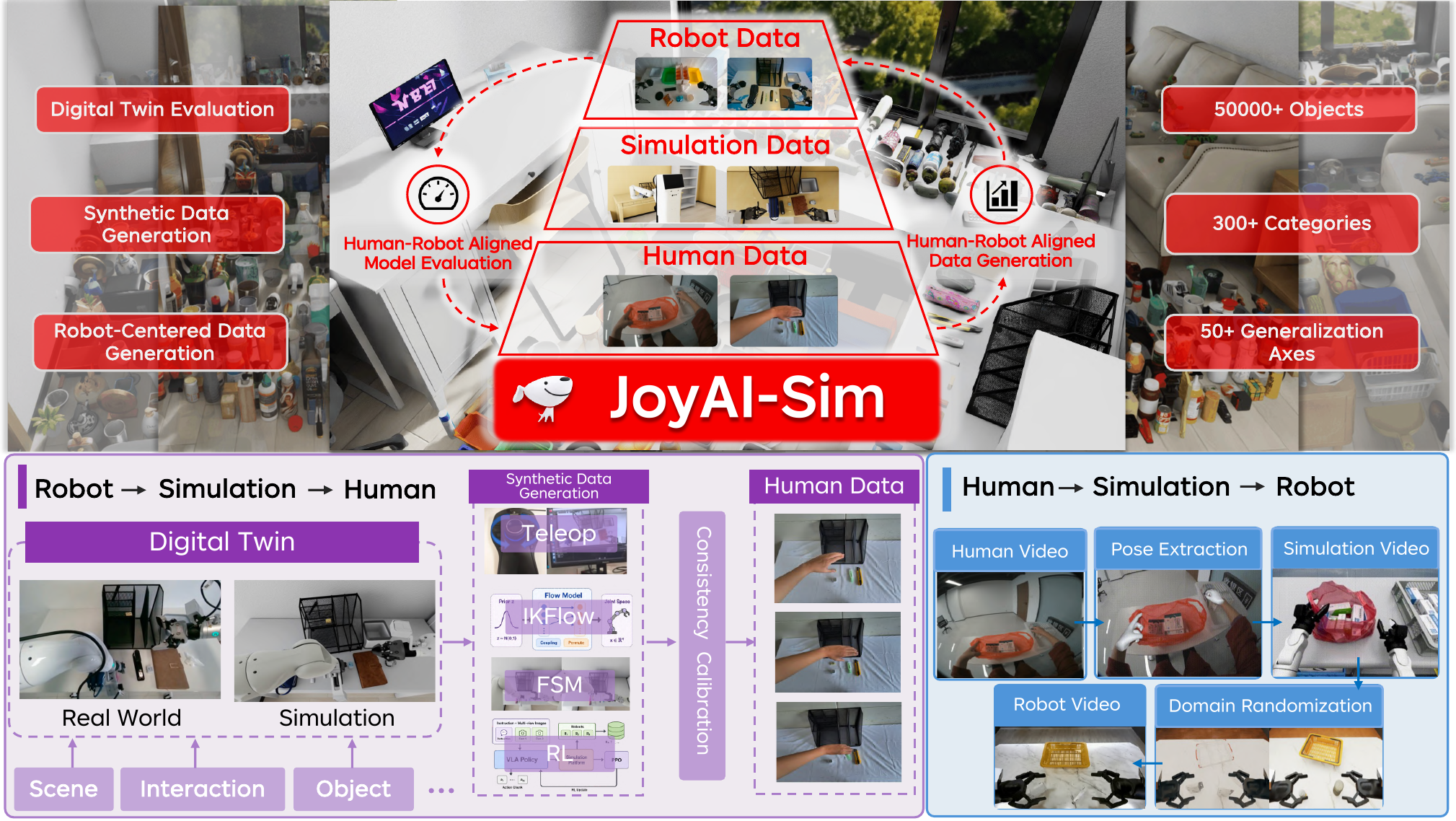}
    \caption{
    \textbf{Overview of JoyAI-Sim.}
    JoyAI-Sim uses the simulator as the central simulation hub to connect robot
    and human data.
    It builds two complementary pathways:
    \textbf{Robot $\rightarrow$ Simulation $\rightarrow$ Human},
    which anchors real-robot tasks in digital twins for human-robot aligned model evaluation,
    and \textbf{Human $\rightarrow$ Simulation $\rightarrow$ Robot},
    which serves as a human-robot aligned data generation pipeline that transforms
    human demonstrations into robot-centered trajectories and robot-view observations
    through simulation.}
    \label{fig:overview}
\end{figure}

We present \textbf{JoyAI-Sim},
a simulation-enabled interconversion toolchain that uses the simulator to align robot data and
human data,
denoted as \textbf{Robot $\rightleftharpoons$ Simulation $\rightleftharpoons$ Human}.
Figure~\ref{fig:overview} summarizes the simulation-centered design.
The first pathway,
\textbf{Robot $\rightarrow$ Simulation $\rightarrow$ Human},
supports \textbf{human-robot aligned model evaluation}.
Real-robot tasks define the deployment target,
calibrated digital twins provide scalable simulation evaluation,
and human embodied feedback is used to inspect the naturalness of simulated
motions.
The second pathway,
\textbf{Human $\rightarrow$ Simulation $\rightarrow$ Robot},
supports \textbf{human-robot aligned data generation}.
Egocentric human demonstrations are lifted into simulation, filtered by
physical feasibility, and converted into robot-centered trajectories and
robot-view observations. Together, the two pathways place simulation at the center of the embodied data pyramid, connecting scarce but deployment-faithful robot data with abundant but embodiment-agnostic human data.

This formulation is broader than the commonly studied real-to-sim-to-real
paradigm, which primarily closes a two-level loop between physical robots and
simulation~\cite{torne2024reconciling,fang2025rebot,wu2025rl,dan2025xsim,
patel2025iker}. At the level of data flow, real-to-sim-to-real mainly covers the
Robot $\rightleftharpoons$ Simulation loop: real-robot scenes or
demonstrations are reconstructed in simulation, used for evaluation or data
generation, and transferred back to the robot. JoyAI-Sim instead extends this
loop into a three-level Robot $\rightleftharpoons$ Simulation
$\rightleftharpoons$ Human framework. Compared with the two-modality
Robot $\rightleftharpoons$ Simulation setting, this formulation
explicitly introduces human data as an additional source of
embodied observations, demonstrations, and feedback. This better matches the
current embodied-data pyramid, where robot data is deployment-faithful but
scarce, simulation data is scalable and physically inspectable, and human data
is abundant but not directly robot-executable.

In the \textbf{Robot $\rightarrow$ Simulation $\rightarrow$ Human} pathway, we use the simulator to support human-robot aligned model evaluation by turning real-robot tasks into scalable simulation evaluations and human-inspectable trajectories.
We begin from standardized real-robot household tidy-up tasks, which provide the physical reference for scene layout, object categories, reset conditions, language-conditioned task semantics, and success criteria.
We then
reconstruct the same scenarios inside the simulator,
which is built on NVIDIA Isaac Sim,
as calibrated digital twins,
enabling candidate policies to be evaluated at scale before costly
real-robot trials.
However,
physically executable trajectories are not always natural or useful for policy
learning.
To assess trajectory quality,
we project simulated robot trajectories into human-hand space for embodied inspection,
enabling human feedback for filtering and improving synthesized data.
This projection further produces aligned human-robot demonstrations as a by-product: each simulated episode can be exported as both robot-centered trajectories and corresponding human-form demonstrations. 
Such aligned human-robot data can support policy-training settings that benefit from paired human- and robot-form demonstrations~\cite{zheng2026egoscale}.

The \textbf{Human $\rightarrow$ Simulation $\rightarrow$ Robot} pathway targets human-robot aligned data generation.
It uses the simulator to turn abundant human demonstrations into physically feasible robot-centered training data.
Instead of directly mapping human hand motions to a robot embodiment, JoyAI-Sim first
recovers human motion and reconstructs the surrounding task scene into a
sim-ready representation. The simulator then retargets the motion under
robot kinematic limits, collision constraints, contact feasibility, and task
geometry. Validated rollouts are rendered from robot viewpoints and can be
augmented through domain randomization and reality augmentation~\cite{
tobin2017domain,peng2018sim2real,alhaija2025cosmostransfer}. This pathway
transforms low-cost human egocentric videos into a scalable and reusable source of
robot-centered trajectories, annotations, and visual observations while
preserving physically meaningful structure.

We package the conversion, simulation, rendering, and realism-augmentation
modules as reusable services on JD Cloud for robot data production and
evaluation.
JoyAI-Sim further integrates with NVIDIA Isaac Lab Arena for unified and
scalable benchmark construction and policy evaluation.

Our contributions are summarized as follows:

\begin{enumerate}
    \item A \textbf{Robot $\rightarrow$ Simulation $\rightarrow$ Human}
    pathway for \textbf{human-robot aligned model evaluation}.
    JoyAI-Sim starts from real-robot long-horizon household tasks and
    reconstructs them as calibrated digital twins for scalable evaluation.
    By exporting each simulated episode into both robot-centered and human-form
    representations,
    JoyAI-Sim supports embodied trajectory inspection,
    while human embodied feedback provides naturalness criteria for inspecting
    and filtering generated trajectories.

    \item A \textbf{Human $\rightarrow$ Simulation $\rightarrow$ Robot}
    pathway for \textbf{human-robot aligned data generation}.
    JoyAI-Sim converts egocentric human demonstrations into sim-ready
    motion and scene representations, checks them in the simulator under robot physical
    constraints, and exports executable robot-centered trajectories and
    robot-view observations. This pathway uses the simulator as a physical
    consistency layer between abundant human videos and scarce robot training
    data, enabling large-scale human demonstrations to be transformed into
    robot-centered training data.

    \item \textbf{Scalable simulation infrastructure.} We deploy JoyAI-Sim on JD Cloud and integrate it with NVIDIA Isaac Lab Arena~\cite{isaaclab-arena2025} for scalable data generation, policy evaluation, and one-to-$N$ environment expansion.
\end{enumerate}

\section{Related Work}
\label{sec:relatedwork}

\paragraph{Robotic Manipulation Benchmarks.}
Real-robot benchmarks provide the most direct evidence of deployability,
because physical trials expose sensing noise,
control latency,
contact uncertainty,
hardware constraints,
and safety requirements.
Large-scale systems and datasets such as RT-1~\cite{brohan2022rt},
RT-2~\cite{brohan2023rt},
Open X-Embodiment~\cite{oneill2024openx},
and DROID~\cite{khazatsky2024droid} have made real-world generalization a central evaluation target.
Recent benchmarks such as RoboArena~\cite{atreya2025roboarena},
RoboChallenge~\cite{yakefu2025robochallenge},
and ManipArena~\cite{sun2026maniparena} further standardize physical evaluation through shared platforms,
fixed protocols,
or distributed evaluators.
However,
real-robot evaluation is expensive,
low-throughput,
and sensitive to reset conditions,
illumination,
object states,
and hardware variation.
Simulation benchmarks such as RLBench~\cite{james2019rlbench},
LIBERO~\cite{liu2023libero},
RoboCasa~\cite{nasiriany2024robocasa},
RoboTwin~\cite{mu2025robotwin},
and RoboTwin 2.0~\cite{chen2025robotwin} provide scalable and reproducible evaluation,
but their scenes are often manually designed and only loosely tied to real deployment instances.
JoyAI-Sim follows the complementary route introduced in Sec.~\ref{sec:introduction}: real robots define the task semantics and success criteria, while a real-world-aligned simulation layer reproduces the assets, robot embodiment, interaction loop, and task predicates for controlled evaluation.


\paragraph{Robot $\rightarrow$ Simulation $\rightarrow$ Human.}
Digital-twin construction is the core technical form of Robot $\rightarrow$ Simulation:
it turns a physical robot scene into a controllable simulator that preserves the task,
the objects,
and the robot interaction conditions.
General-purpose simulators such as Isaac Gym~\cite{makoviychuk2021isaac},
SAPIEN~\cite{xiang2020sapien},
Habitat~\cite{savva2019habitat},
AI2-THOR~\cite{kolve2017ai2thor},
and iGibson~\cite{shen2021igibson} provide the physics and rendering substrate.
On top of these systems,
scene reconstruction methods build editable digital twins from real deployments:
RialTo~\cite{torne2024reconciling} scans a workspace into a policy-training twin,
Robo-GS~\cite{11128786} combines Gaussian Splatting and mesh assets for articulated reconstruction,
and GSWorld~\cite{jiang2025gsworld} and GaussGym~\cite{escontrela2025gaussgym} use Gaussian-Splatting-based rendering to reduce visual mismatch.
These works show that Robot $\rightarrow$ Simulation can support both scalable
policy training and real-world policy evaluation,
including cases where simulated rankings correlate with physical
performance~\cite{li2024simpler,abouchakra2025realissim}.
Another related direction evaluates simulated robot motions through external
critics:
CRISP~\cite{lim2026robotsinnercriticselfrefinement} employs vision-language
models (VLMs) to assess action appropriateness and iteratively refine behaviors
through a generate--evaluate--replan loop in simulation.
However,
because VLMs primarily rely on semantic visual understanding rather than
embodied physical intuition,
their assessments may be insufficient to evaluate motion naturalness and physical feasibility.
To address this gap,
JoyAI-Sim introduces a Simulation $\rightarrow$ Human paradigm,
where synthesized robot trajectories are transferred into human-executable
spaces for direct inspection,
allowing humans to identify unnatural or impractical behaviors through embodied
feedback and physical intuition. JoyAI-Sim differs by building household digital twins for two explicit purposes: first, to align with real-robot evaluation through scene, asset, embodiment, action, and predicate alignment; second,
to synthesize policy-training data that can be further inspected through
Simulation $\rightarrow$ Human feedback and exported as aligned robot-human
trajectories.

\paragraph{Human $\rightarrow$ Simulation $\rightarrow$ Robot.}
Human $\rightarrow$ Simulation $\rightarrow$ Robot methods use human behavior
as a scalable source of manipulation priors,
and introduce simulation to bridge the gap between human demonstrations and
robot execution.
Large-scale egocentric datasets such as Ego4D~\cite{grauman2022ego4d} and
EgoLive~\cite{li2026egolive} capture diverse everyday hand-object interactions,
while hand-recovery methods such as HaMeR~\cite{pavlakos2024hamer} estimate
task-relevant wrist motion,
hand pose,
and manipulation phases from monocular videos.
However,
these outputs remain human-centered observations rather than robot
demonstrations,
because human hands and robot end-effectors differ in kinematics,
contact geometry,
compliance,
and feasible force profiles.

Recent work addresses parts of this gap with simulation,
view alignment,
or task-centric rewards.
EgoHumanoid~\cite{shi2026egohumanoid} studies view and action alignment from
egocentric demonstrations,
X-Sim~\cite{dan2025xsim} and IKER~\cite{patel2025iker} use human videos with
object- or keypoint-centric rewards,
and DexMan~\cite{hsieh2025dexman} converts human and generated videos into
dexterous manipulation skills in simulation.
JoyAI-Sim follows this line but organizes the full pathway under one
simulation-centered data-production toolchain:
human video parsing,
editable simulation instantiation,
robot feasibility checking,
robot-view rendering,
and downstream data augmentation.

\paragraph{Synthetic Data Generation.}
Simulation-based data synthesis significantly reduces the cost of collecting
robot demonstrations and has become a key approach for scaling manipulation
datasets~\cite{mandlekar2023mimicgen,nasiriany2024robocasa,
chen2025robotwin,wang2023gensim,tobin2017domain}.
One line of work relies on teleoperation systems to collect human
demonstrations,
with some efforts focusing on data collection in high-fidelity simulation
environments,
enabling low-cost acquisition of robot manipulation trajectories~\cite{
li2025teleopbenchsimulatorcentricbenchmarkdualarm,
mandlekar2018roboturkcrowdsourcingplatformrobotic}.
Beyond human demonstrations,
expert policies based on finite state machines (FSMs)~\cite{
iovino2024comparisonbehaviortreesfinite} can directly generate trajectories
from task structures,
providing scalable supervision when demonstrations are scarce.
Another direction augments a small set of feasible trajectories through
generative inverse kinematics;
for example,
IKFlow~\cite{ames2022ikflowgeneratingdiverseinverse} and related
methods~\cite{zhang2026ikdiffuserdiffusionbasedgenerativeinverse} generate
diverse joint-space solutions for the same end-effector pose,
thereby increasing trajectory diversity.
In addition,
reinforcement learning can serve as a data generator,
where reward-driven policies automatically explore environments and collect
both successful and failed trajectories for downstream learning~\cite{
Xu2024RLDGRG,Kanehira2025RLDrivenDG}.
JoyAI-Sim integrates these complementary approaches within digital-twin
environments,
combining teleoperation-based data collection,
FSM-based automatic generation,
IKFlow-based trajectory augmentation,
and reinforcement-learning-driven trajectory mining to construct large-scale
robot simulation datasets.

\section{Robot $\rightarrow$ Simulation $\rightarrow$ Human}
\label{sec:sim_eval}

The Robot $\rightarrow$ Simulation $\rightarrow$ Human pathway starts from real-robot tasks and uses simulation as the central layer for deployment-oriented evaluation and trajectory inspection, with data synthesis serving as an auxiliary source of candidate trajectories.
Real-robot evaluation provides the most faithful deployment signal,
but it is slow,
low-throughput,
and difficult to reproduce under identical object states,
illumination,
and reset conditions.
Following the simulation-based manipulation benchmarks such as SIMPLER~\cite{li2024simpler}, the Robot $\rightarrow$ Simulation step reconstructs real-robot evaluation scenes as calibrated digital twins, so that each task can be converted into a repeatable and perturbable evaluation instance.
Within these digital twins, policies can be screened under controlled variations, such as robot state, object pose, scene layout, illumination, and so on, before final physical validation.
To provide a reliable evaluation substrate, the simulation must consistently and accurately align not only the visible scene but also the robot embodiment, assets, camera configuration, control interface, and task-level success predicates.

Meanwhile, physically executable robot trajectories are not necessarily natural.
A trajectory that succeeds in simulation may still contain awkward approach directions, abrupt phase transitions, or locally feasible but globally suboptimal strategies.
This motivates the Simulation $\rightarrow$ Human step, where simulated robot trajectories are projected into human-hand space for embodied inspection and data alignment.
Related efforts such as EgoInteract~\cite{leonardi2026egointeract} show that egocentric human-hand interactions can be organized in a controllable simulator.
In our setting, this projection preserves the structured interaction process behind the video, including camera motion, hand pose, object state, contact timing, occlusion, and task outcome.
It therefore allows humans to inspect whether a simulated trajectory is natural from an embodied perspective, while also creating one-to-one aligned robot-human data from the same simulated episode.
The $\pi_{0.6}^{*}$ framework~\cite{intelligence2025pi} further highlights the value of failed executions and corrective interventions, which provide supervision for the error states that a policy actually visits during deployment.
Accordingly, robot failures reconstructed in simulation can be converted into human-perspective recovery examples, including failed attempts, corrective motions, recovery behavior, and boundary contact cases.

\begin{figure}[!t]
\centering
\includegraphics[width=\linewidth]{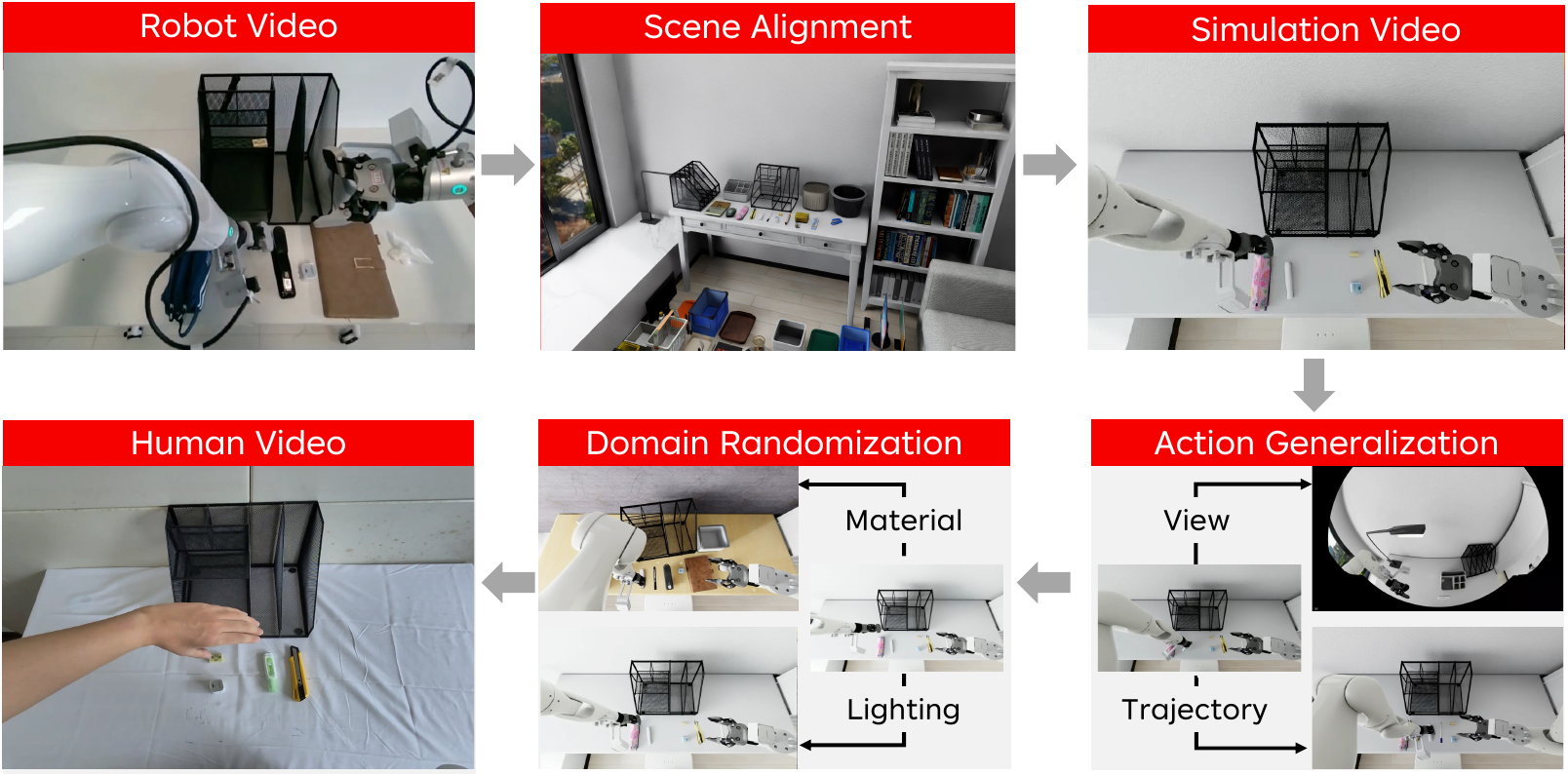}
\caption{\textbf{Robot $\rightarrow$ Simulation $\rightarrow$ Human Toolchain.}
\textbf{Stage~I: Robot $\rightarrow$ Simulation.}
Real-robot household tasks are reconstructed as calibrated digital twins by aligning scenes,
assets,
robot embodiment,
cameras,
actions,
and task predicates,
yielding controllable replicas for scalable evaluation and data synthesis.
\textbf{Stage~II: Simulation $\rightarrow$ Human.}
Generated robot trajectories are projected into human-hand space,
where embodied inspection filters unnatural motions and creates one-to-one
aligned robot-human trajectory pairs.}
\label{fig:robot_simulation_human}
\end{figure}

Together, Robot $\rightarrow$ Simulation constructs a controllable digital-twin task space, while Simulation $\rightarrow$ Human provides embodied naturalness criteria for inspecting, filtering, and aligning simulation-generated trajectories.
Compared with direct Robot $\rightarrow$ Human conversion, this intermediate simulation layer avoids reducing the problem to appearance transfer and instead decomposes limited real-robot data into reusable, expandable, and cross-embodiment-aligned resources.
Figure~\ref{fig:robot_simulation_human} summarizes this two-stage toolchain.

\subsection{Robot $\rightarrow$ Simulation}
\label{sec:robot_to_sim}

\subsubsection{Real-robot Evaluation}
\label{sec:JoyAI-Sim_real}

Real-robot evaluation is the most direct way to measure whether a policy can be
deployed in the physical world.
It exposes sensing noise,
calibration error,
contact uncertainty,
safety behavior,
and long-horizon recovery under real household clutter.
We therefore build our real-robot benchmark as the physical reference for task
semantics,
object-to-target mappings,
scene layouts,
success criteria,
and final deployment-oriented validation.
The benchmark focuses on long-horizon household tidy-up tasks that require
category-level grounding,
precise placement,
sequential execution,
and bimanual manipulation.

\paragraph{Robotic Embodiment Platform.}
The real-world testbed is based on a bimanual humanoid robot platform equipped with two 7-DoF arms, parallel-jaw grippers, a head-mounted RGB-D camera, and two wrist cameras. During each episode, proprioceptive states, end-effector poses, gripper widths, multi-view visual observations, executed actions, timestamps, and task metadata are synchronously logged at 30 Hz. The observation and action interface is kept fixed across all evaluated policies, ensuring that performance differences reflect policy behavior rather than variations in the evaluation setup.

\paragraph{Task Suite.}
Our benchmark contains two long-horizon,
language-conditioned household tidy-up scenarios.
Each scenario is defined by a physical scene,
manipulated objects,
target storage regions,
and object-to-target assignments that are fixed for each episode.

\begin{itemize}
    \item \textbf{Study-Room Tidy-up.}
    The robot clears a study desk by sorting stationery,
    books,
    small daily-use items,
    and trash into their corresponding storage regions.
    This task stresses category-level sorting,
    object-scale variation,
    target-region grounding,
    and long-horizon sequential execution.

    \item \textbf{Living-Room Tidy-up.}
    The robot tidies a cluttered coffee table by sorting tissues,
    toys,
    remote-control items,
    drinks,
    snacks,
    cups,
    and trash into their corresponding containers or support regions.
    This task stresses dense tabletop clutter,
    heterogeneous target containers,
    visually diverse object categories,
    and fine-grained language grounding.
\end{itemize}

\paragraph{Evaluation Protocol and Metrics.}
For each task, policy, and evaluation condition, trials are initialized from a predefined reset list, so that the same initial configurations can be reused across policies.
Object poses are randomized within the robot's reachable workspace and the
camera field of view,
while target containers,
shelves,
storage boxes,
and bins are either fixed or sampled from predefined layouts depending on the
evaluation split.
The policy then executes autonomously without human correction.
An episode terminates when all required sub-tasks are completed,
the maximum horizon $T_{\max}$ seconds or $H_{\max}$ control steps is reached,
or a safety termination is triggered.

A full task is successful only if all required objects are placed into their
correct target regions within the episode horizon,
and no safety termination occurs.
We also report sub-task completion rate,
grasp success rate,
and placement accuracy.
Sub-task completion measures the fraction of required atomic assignments
completed in an episode.
Grasp success measures whether the robot lifts and stabilizes the intended
object after a grasp attempt.
Placement accuracy measures whether the final object state lies inside the
assigned target region or compartment under the task-specific tolerance
$\epsilon$.

\paragraph{Generalization Axes.}
Instead of treating robustness as a single aggregate score,
the benchmark evaluates controlled generalization dimensions while keeping task
semantics and success criteria fixed.
Representative visual generalization conditions are illustrated in Appendix~\ref{app:real-robot-generalization-observations}
using head-camera observations collected from the real robot. Each condition changes one non-semantic factor while preserving the task semantics and each object's assigned target configuration in the benchmark.

\FloatBarrier


\begin{figure*}[t]
\centering

\begin{subfigure}[htbp]{0.49\linewidth}
    \centering
    \includegraphics[width=\linewidth]{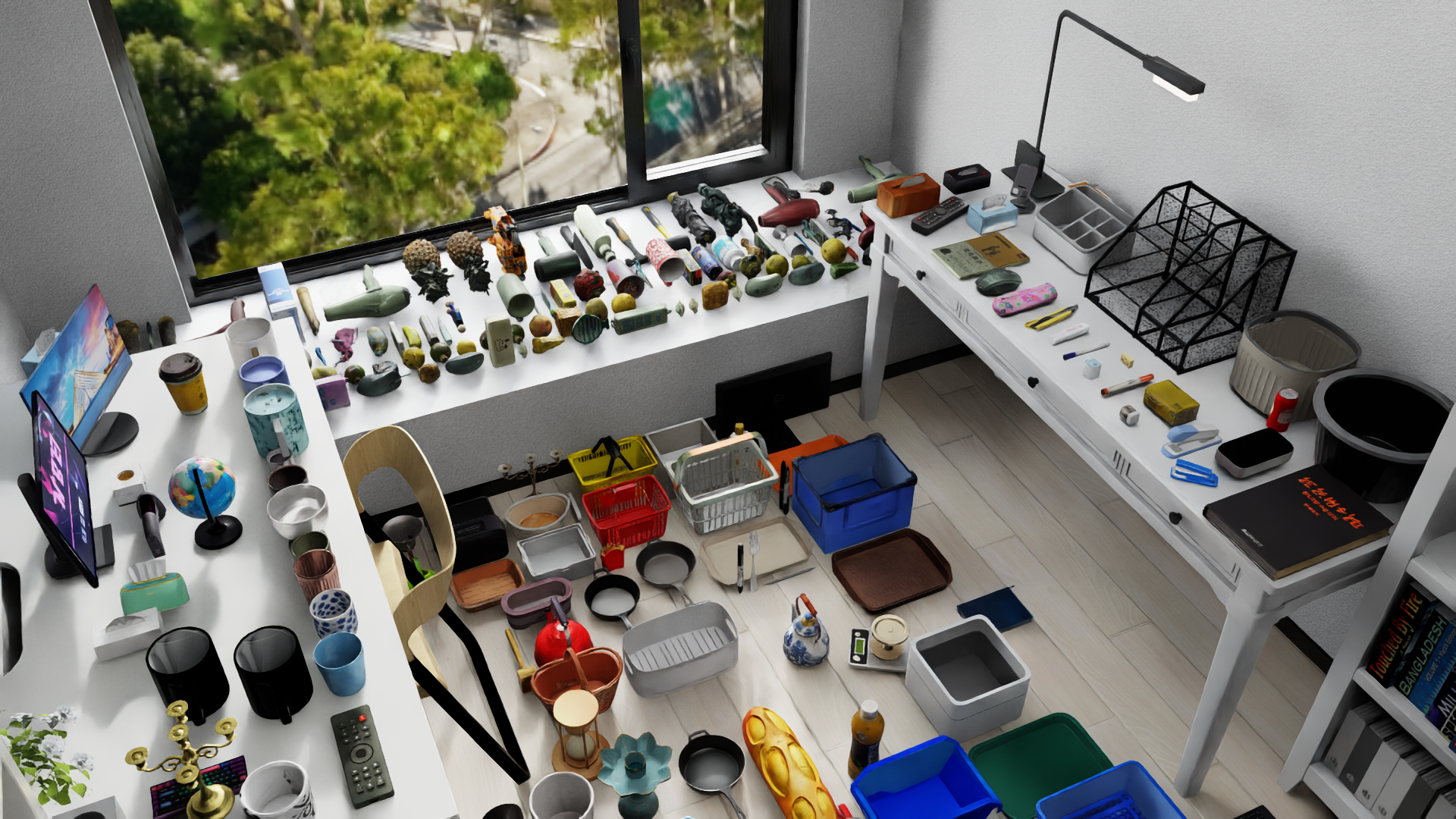}
    \caption{Study Room}
    \label{fig:real2sim_pipeline_c}
\end{subfigure}
\hfill
\begin{subfigure}[htbp]{0.49\linewidth}
    \centering
    \includegraphics[width=\linewidth]{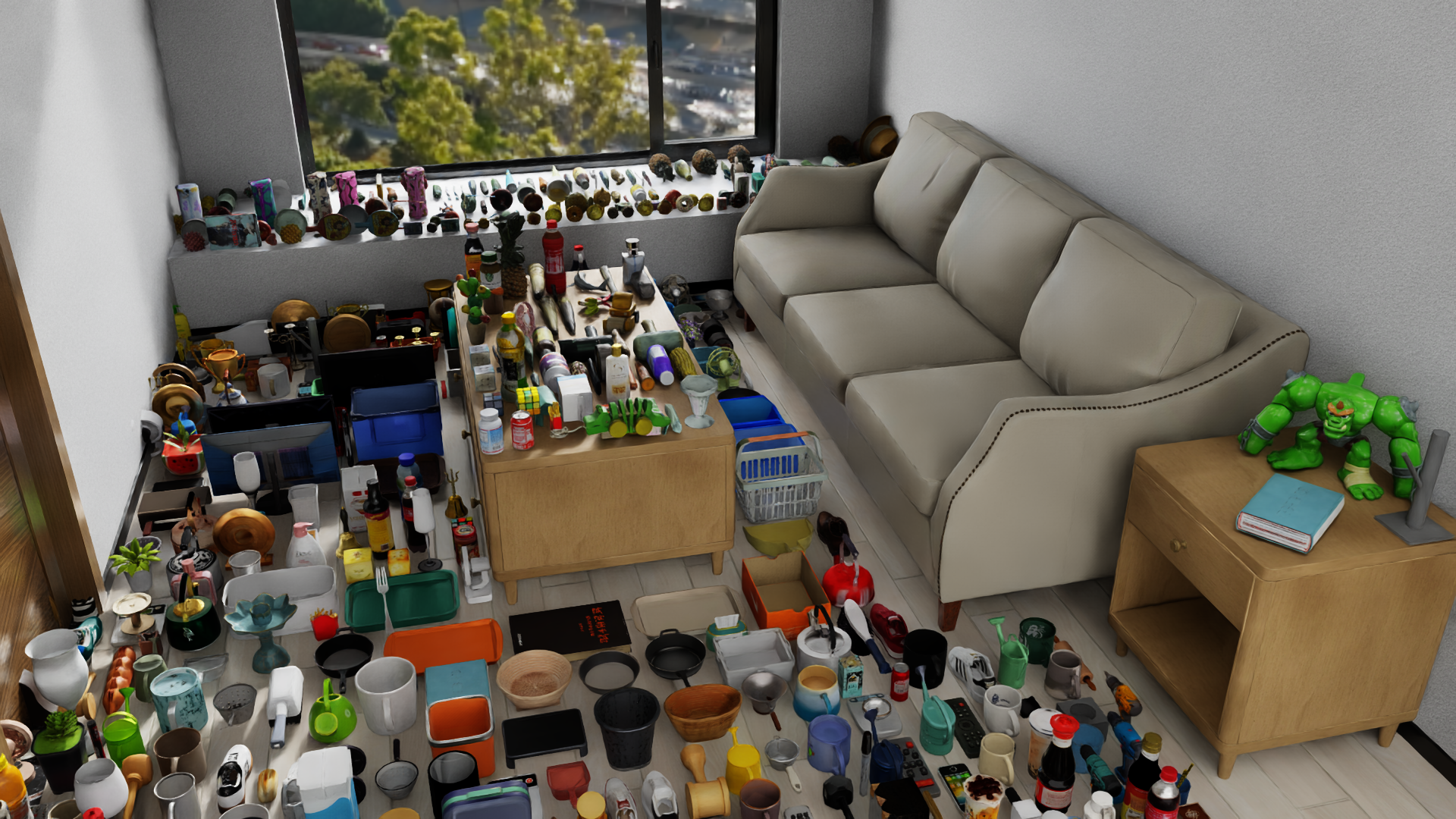}
    \caption{Living Room}
    \label{fig:real2sim_pipeline_d}
\end{subfigure}

\caption{\textbf{Diverse Assets in the Simulator.}}
\label{fig:real2sim_pipeline-2}
\end{figure*}

\subsubsection{Scene Alignment}
\label{sec:JoyAI-Sim-scene-environment-alignment}

\paragraph{Sim-Ready Data Preparation.}
We build the simulator asset library through two complementary sources: collecting assets from public datasets, normalizing them into a common Isaac Sim format, and reconstructing task-specific assets in 3D when needed.
Since some assets were originally authored for Isaac Sim~4.1 and 4.5,
we convert and import them into Isaac Sim~5.1.
After import,
we check loading,
collision geometry,
and material integrity to avoid corruption introduced by cross-version
conversion.
After reorganizing the asset library for household scenes, we obtain $300{+}$ fine-grained categories comprising 53{,}661 asset instances, as shown in Table~\ref{tab:assets}.
Additional asset distribution visualizations are provided in
Appendix~\ref{app:sim-ready-assets}.

\paragraph{Sim-Ready Scene Preparation.}
Based on this asset library, we construct two household scenes with diverse assets for the real-robot evaluation setup:
the study room in Figure~\ref{fig:real2sim_pipeline_c},
and the living room in Figure~\ref{fig:real2sim_pipeline_d}.
For each scene,
we align the simulator with the real workspace in appearance,
geometry,
object placement,
and task-relevant spatial layout.
We use 3D Gaussian Splatting and image-to-3D generation to recover high-fidelity
scene and object assets,
and then enrich the manipulated-object set with library assets.
The resulting scenes preserve the physical structure needed for real-robot
evaluation,
while also supporting controlled variations over object category,
object placement,
illumination,
and texture for large-scale policy-training data generation.

\begin{figure}[!htbp]
\centering
\includegraphics[width=\linewidth]{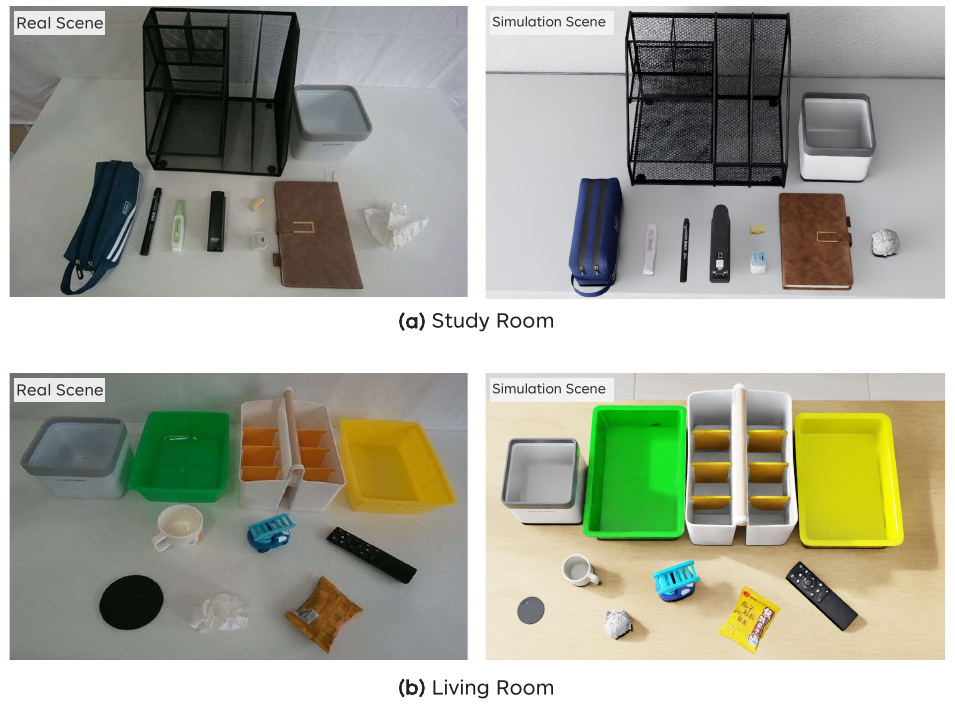}
\caption{
\textbf{Paired Real-Robot and Simulation Scenes.}
Each pair preserves task-relevant containers, objects, spatial arrangements,
and robot states between the real-robot setup and its corresponding
simulation scene.}
\label{fig:robot_simulation_scene_alignment}
\end{figure}

\paragraph{Real-Simulation Scene Alignment.}
To make robot observations comparable with simulation observations,
we construct paired robot and simulation scenes for both tidy-up tasks.
As shown in Figure~\ref{fig:robot_simulation_scene_alignment},
each simulated scene preserves the task-relevant containers, object categories, spatial arrangement, and robot state observed in the real-robot workspace.
This pairing keeps the language goal,
object-to-target mappings,
and success criteria consistent between the real robot and the simulator,
while allowing controlled changes in appearance,
illumination,
layout,
and robot state for scalable evaluation.

\subsubsection{Interaction Alignment}
\label{sec:JoyAI-Sim-embodiment-alignment}
\label{sec:joysim-embodiment-alignment}

JoyAI-Sim runs digital twins in simulation, where interaction alignment is implemented through two coupled tracks: embodiment alignment and action alignment. Embodiment alignment ensures that the simulated robot platform matches the physical robot in kinematics, initial state, sensing geometry, and actuator dynamics.
Action alignment allows the real robot and digital twin to share the same control schema, trajectory record-and-replay stream, and episode lifecycle protocol.
Together,
these two tracks allow the digital twin to serve as a simulation-first
measurement tool,
rather than merely a visually similar replica.
Table~\ref{tab:embodiment-interaction-alignment} summarizes the embodiment- and action-level design choices used for this alignment.

\begin{table}[htbp]
  \centering
    \caption{
    \textbf{Key Elements of Simulator Interaction Alignment.}
    Interaction alignment combines embodiment-level consistency in kinematics, initial state, sensing geometry, and actuator dynamics with action-level consistency in control, rollout replay, and episode lifecycle protocol.
    }
  \label{tab:embodiment-interaction-alignment}
  \small
  \setlength{\tabcolsep}{3.0pt}
  \renewcommand{\arraystretch}{1.04}
  \begin{tabularx}{\linewidth}{@{}>{\centering\arraybackslash}p{0.13\linewidth}>{\centering\arraybackslash}p{0.13\linewidth}>{\centering\arraybackslash}X@{}}
    \toprule
    \textbf{Track} & \textbf{Element} & \textbf{Aligned design and evaluation role} \\
    \midrule
    \multirow{4}{=}{Embodiment}
      & Kinematics
      & SDK-consistent URDF/USD models for shared IK, indices, and end-effector frames \\
    & Initial state
      & Encoder-based ready pose with settling for stable, comparable starts \\
    & Sensing
      & Calibrated camera and hand--eye geometry for shared visual observations \\
    & Dynamics
      & Subsystem stiffness/damping for calibrated reach, compliance, and settling \\
    \midrule
    \multirow{3}{=}{Action}
      & Control
      & Unified Data Distribution Service schema for action semantics without remapping \\
    & Rollout
      & 30\,Hz Parquet record/replay for the same command sequence $\mathbf{u}_{0:T}$ \\
    & Protocol
      & Prepare/finalize/reset commands for repeatable episode boundaries \\
    \bottomrule
  \end{tabularx}
\end{table}


\paragraph{Embodiment Alignment.}
The simulated robot platform uses SDK-consistent URDF and USD robot models.
The URDF supports inverse kinematics,
teleoperation,
and trajectory replay,
while the USD articulation supports Isaac~Sim physics,
collision modeling,
and camera attachment.
The two assets share the same joint names,
joint ordering,
link hierarchy,
and end-effector frames as the physical robot SDK,
allowing inverse kinematics,
control indices,
and end-effector poses to carry the same meaning in real and simulated
executions.
At reset,
the robot is restored to an encoder-based ready pose,
followed by a short settling phase before policy execution,
so each episode starts from a stable and comparable state.
The head and wrist cameras use calibrated camera intrinsics and hand--eye
extrinsics,
allowing real and simulated logs to share the same visual observation format
for the VLA pipeline.
Actuator behavior is calibrated with subsystem-specific stiffness and damping
parameters for the arms,
torso,
head,
grippers,
and passive finger links,
reducing mismatch in reaching,
contact compliance,
and settling behavior.

\paragraph{Action Alignment.}
Policies,
teleoperators,
and replay scripts use a unified Data Distribution Service (DDS) message schema
on both the simulator and the real robot.
Switching between the two only changes the DDS endpoint,
while preserving the same action semantics without remapping,
rescaling,
or clipping the action space.
Teleoperation and policy rollouts are both recorded as timestamped Parquet streams at
30\,Hz,
including proprioceptive states,
end-effector poses,
gripper widths,
and multi-view robot observations.
The same command sequence $\mathbf{u}_{0:T}$ can then be replayed in the simulator
through the shared control interface.
For batch evaluation,
the prepare,
finalize,
and reset lifecycle commands standardize recorder startup,
log finalization,
scene restoration,
and episode boundaries,
so policy checkpoints can be compared under repeatable evaluation conditions across repeated runs.



\subsubsection{Simulation Evaluation}
\label{sec:JoyAI-Sim-simulation-evaluation}

Simulation evaluation operationalizes the preceding alignment steps.
The real-robot benchmark explicitly defines what should be evaluated: task semantics, object-to-target mappings, reset conditions, episode horizons, and success criteria.
Scene alignment determines where the evaluation is performed by reconstructing the study-room and living-room tasks as digital twins.
Interaction alignment determines how the evaluation is executed by matching the robot embodiment, sensing layout, control interface, trajectory replay, and episode lifecycle between hardware and simulation.
With these components aligned, the simulator serves not only as a rendering and physics backend, but also as a controlled layer for evaluating deployment-oriented policies before physical trials.

The role of this evaluation layer is complementary to real-robot testing.
Real-robot trials remain the final measure of deployable performance, because they expose contact uncertainty, sensing artifacts, calibration error, and hardware-level safety behavior.
However, physical trials are expensive and make it difficult to isolate a single cause of failure.
We use the aligned digital twins to test whether a policy is sensitive to specific non-semantic factors, including robot initialization, object placement, object instance, background appearance, illumination, and instruction wording.
To isolate each factor, we replay the same task from matched reset conditions and vary only one factor at a time while preserving the intended task goal.

For each task, JoyAI-Sim uses the simulator to generate evaluation episodes by systematically varying a predefined set of generalization axes.
Robot-state perturbations test whether the policy depends on a narrow base pose or end-effector ready pose.
Object-layout perturbations reveal spatial reasoning errors, collision-prone behavior, and wrong placement under clutter or occlusion.
Object-instance perturbations evaluate whether grasping and placement remain stable when object geometry or physical properties change.
Background, surface, and illumination variations diagnose visual overfitting in perception and pose estimation.
Instruction paraphrases test whether the policy grounds objects, containers, shelf layers, and target regions by task semantics rather than by memorized wording.
Table~\ref{tab:sim-generalization-axes} summarizes these axes, their controlled variations, and the corresponding diagnostic signals.

\begin{table}[tb]
    \centering
    \footnotesize
    \setlength{\tabcolsep}{4.2pt}
    \renewcommand{\arraystretch}{1.18}
    \caption{
    \textbf{Generalization Axes for Simulation-Based Policy Evaluation.}
    JoyAI-Sim generates evaluation episodes by varying predefined axes, including robot state, object layout, object instance, background, surface, illumination, and instruction phrasing.
    }
    \label{tab:sim-generalization-axes}
    \begin{tabularx}{\linewidth}{
        @{}
        >{\centering\arraybackslash}m{0.20\linewidth}
        >{\centering\arraybackslash}p{0.15\linewidth}
        >{\centering\arraybackslash}X
        >{\centering\arraybackslash}X
        @{}
    }
        \toprule
        Example & Axis Categories & Controlled Variation & Diagnostic Signal \\
        \midrule
        \includegraphics[width=0.96\linewidth]{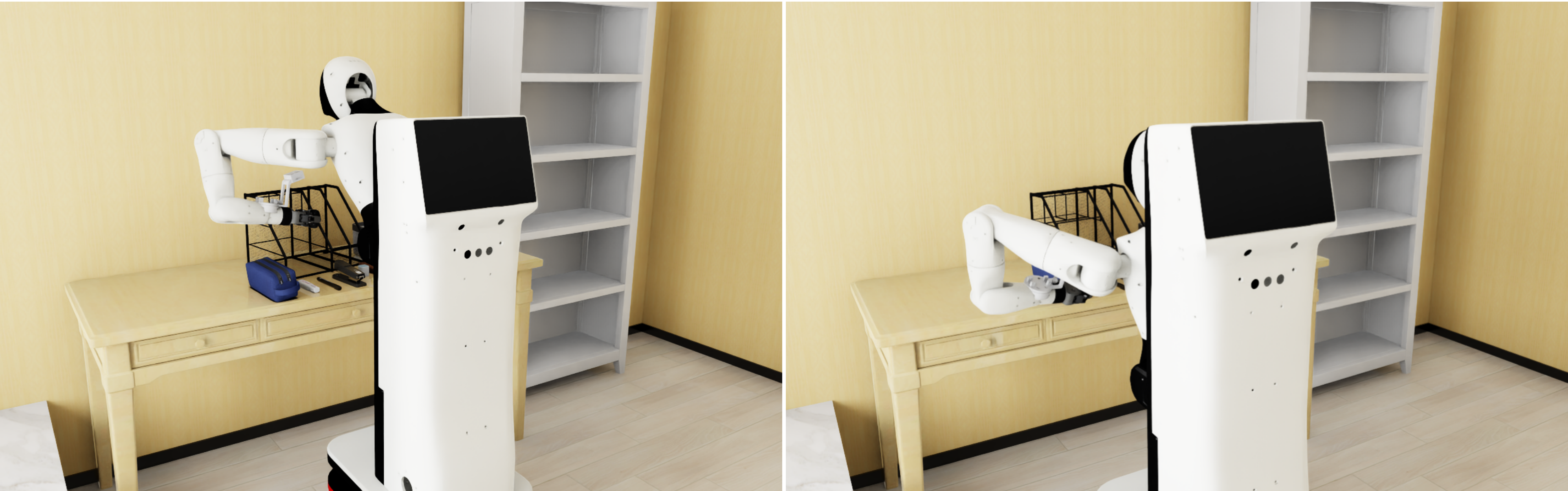}
        & Robot State
        & Base offset; end-effector ready pose
        & Approach and pre-grasp sensitivity \\
        \midrule
        \includegraphics[width=0.96\linewidth]{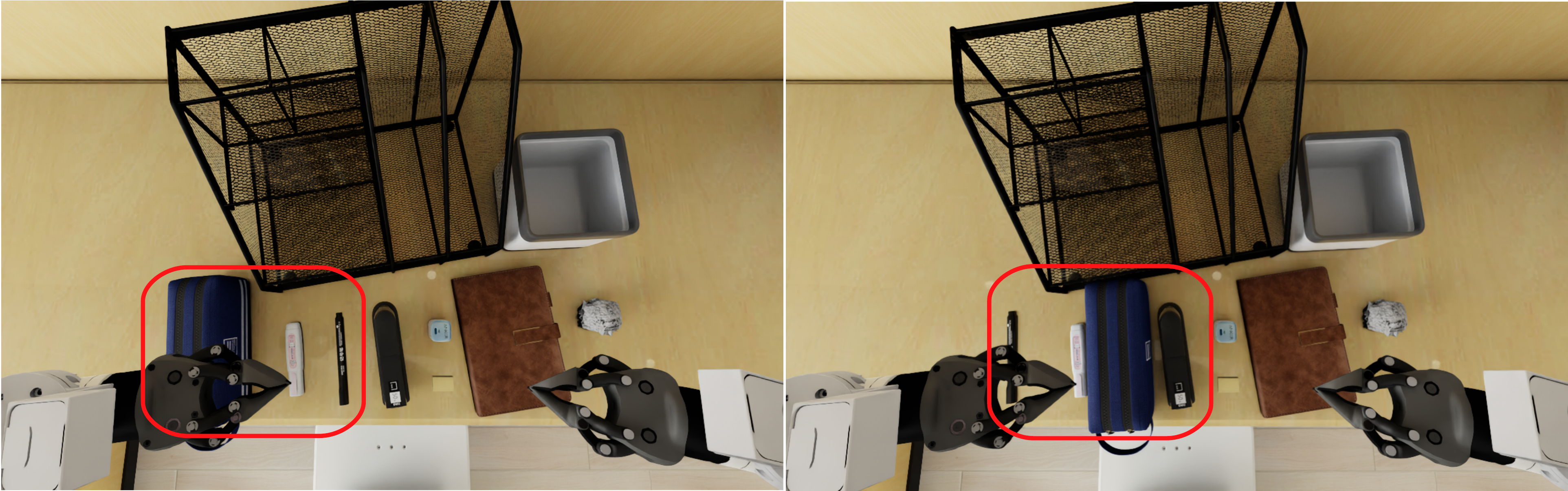}
        & Object Layout
        & Target, distractor, and receptacle pose; clutter and occlusion
        & Spatial error, collision, or wrong placement \\
        \midrule
        \includegraphics[width=0.96\linewidth]{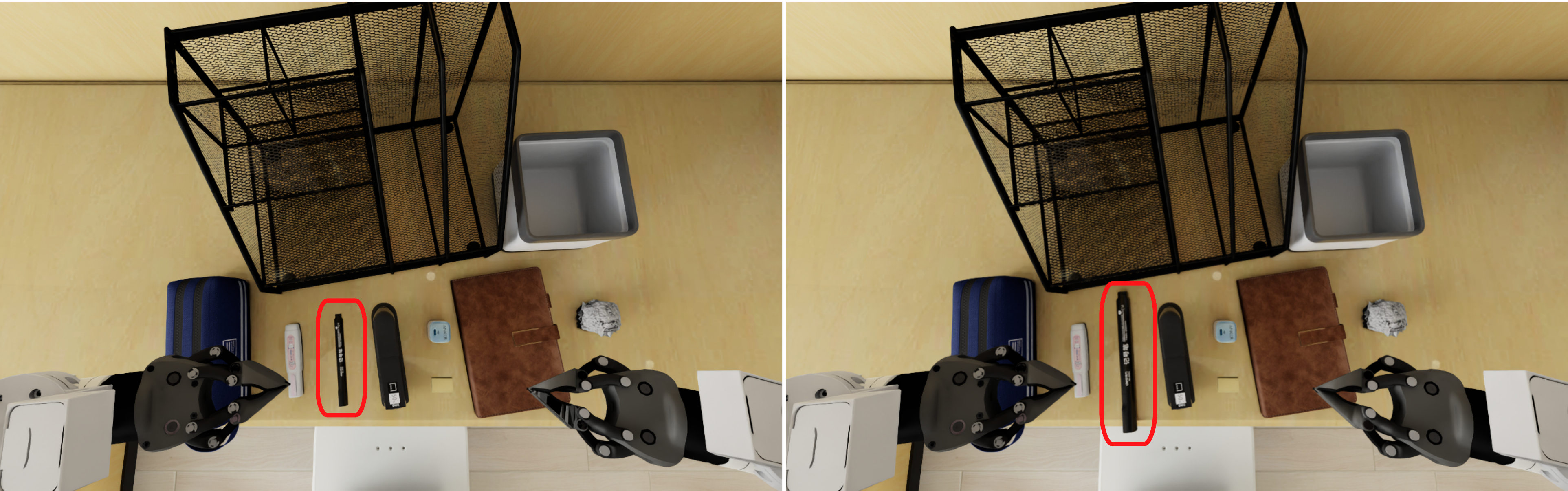}
        & Object Instance
        & Shape, size, material, mass, center of mass, and deformability
        & Grasp mismatch, unstable lift, or object drop \\
        \midrule
        \includegraphics[width=0.96\linewidth]{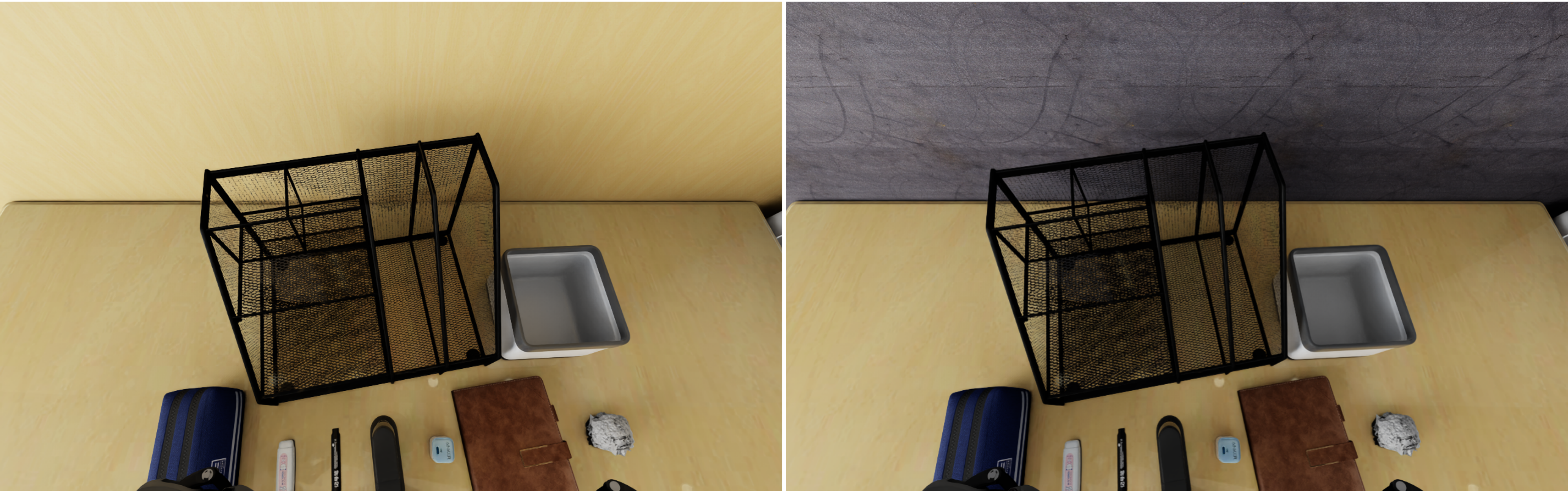}
        & Background / Surface
        & Texture, color, reflectance, and non-task visual distractors
        & Pose-estimation drift under visual changes \\
        \midrule
        \includegraphics[width=0.96\linewidth]{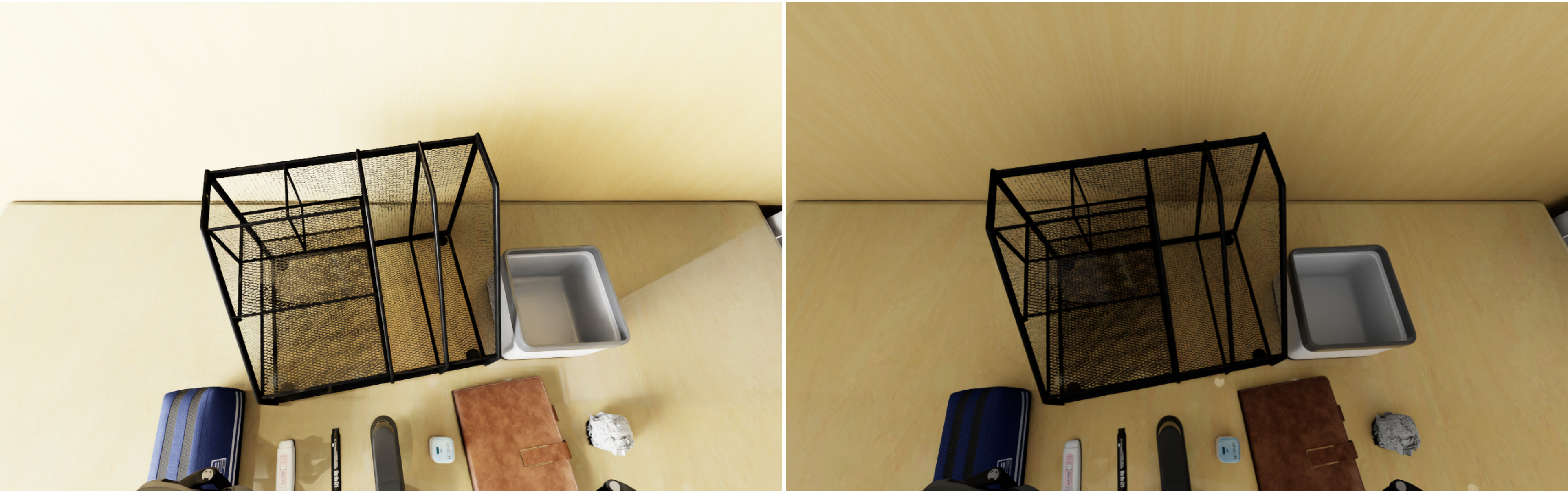}
        & Illumination
        & Brightness, illumination direction, shadows, and dim or hard illumination
        & Detection, pose, or placement error under photometric shift \\
        \midrule
        Put the black pen in the mesh rack. \textbf{vs.} Place the black pen into the mesh organizer.
        & Task Instruction
        & Equivalent references to object, slot, layer, or region
        & Wrong-object or wrong-goal grounding \\
        \bottomrule
    \end{tabularx}
\end{table}

The evaluation target is object-centric.
For each task-relevant object, the simulator records the final stable pose and checks it against the assigned goal region defined by the task.
For placement and tidy-up tasks, the pose is evaluated in the coordinate frame of the target container, shelf, bin, coaster, or support region.
This prevents a global-position match from being treated as success when the object must be placed into a specific compartment, layer, or container.
When a full 6-DoF match is unnecessary, only the relevant pose components are checked, such as planar position, support height, container membership, or compartment assignment for the target object.

An episode is counted as successful only when all required objects reach their assigned target states within the horizon, and no safety termination is triggered.
In addition to the final success or failure result, the evaluator records per-object completion, grasp outcome, placement error, collision events, timeout, safety-stop status, and success rate under each perturbation axis.
These records turn a single task-level score into a failure profile: layout-sensitive failures, illumination-sensitive failures, grounding errors, and grasp-instability failures can be separated and addressed through targeted data generation, scene randomization, policy refinement, or selective real-robot validation.

This protocol closes the Robot $\rightarrow$ Simulation part of JoyAI-Sim.
The real robot anchors the benchmark in deployment-relevant tasks, scene alignment establishes the digital-twin environment, interaction alignment preserves the execution interface, and simulation-based evaluation turns the aligned system into a scalable measurement instrument.
The resulting diagnostic profiles support simulation-first checkpoint screening before real-robot deployment, and also provide the basis for the following Simulation $\rightarrow$ Human stage, where simulated trajectories are inspected with human naturalness criteria before being used as training data.

\subsection{Simulation $\rightarrow$ Human}
\label{sec:simulation_to_human}

\subsubsection{Synthetic Data Generation}
\label{sec:simulation-synthetic-data}

The Simulation $\rightarrow$ Human stage starts with candidate robot trajectories generated in simulation, which are then projected into human-hand space for embodied inspection and robot-human data alignment.
To improve their coverage and diversity, we develop four complementary simulation-based data synthesis and augmentation pipelines.
First,
a virtual teleoperation framework is employed,
where an operator uses a six-degree-of-freedom input device to remotely control
a robot in the simulator and collect demonstration data.
Second,
a finite-state-machine (FSM)-based pipeline enables rule-driven automatic data
generation.
Third,
IKFlow is leveraged to efficiently augment existing trajectories in the joint
space.
Finally,
reinforcement learning (RL) is used to train general manipulation policies that automatically generate diverse task data.
As summarized in Table~\ref{tab:data_augmentation_compare},
the four approaches exhibit different characteristics in terms of demonstration
requirements,
trajectory acceptance,
and wall-clock synthesis speed for a 10-second simulated episode.

\begin{table}[!htbp]
\centering
\small
\caption{
\textbf{Comparison of Simulator-Based Synthetic Data Generation Methods.}
Teleoperation, FSM-based generation, IKFlow-based augmentation, and RL-based generation are compared in terms of demonstration requirements, trajectory acceptance, and synthesis speed for a 10-second simulated episode.
}
\begin{tabular}{lccc}
\toprule
Method
& Seed Demonstrations Required
& Acceptance Rate
& Synthesis Speed (s/episode) \\
\midrule
Teleoperation & None & $\sim 80\%$ & $\sim 380$ \\
FSM     & None & $\sim 30\%$ & $\sim 300$ \\
IKFlow  & Few  & $\sim 70\%$ & $\sim 15$  \\
RL      & Many & $\sim 90\%$ & $\sim 10$  \\
\bottomrule
\end{tabular}
\label{tab:data_augmentation_compare}
\end{table}

\begin{figure*}[!htbp]
\centering

\begin{subfigure}[t]{0.98\textwidth}
    \centering
    \includegraphics[width=\linewidth]{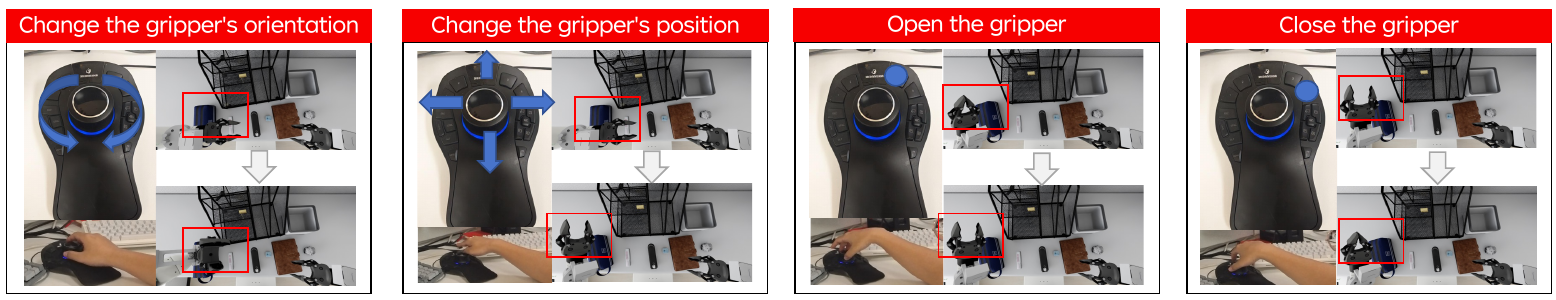}
    \caption{Teleoperation in the Simulator}
    \label{fig:tele}
\end{subfigure}

\vspace{2mm}

\begin{subfigure}[t]{0.98\textwidth}
    \centering
    \includegraphics[width=\linewidth]{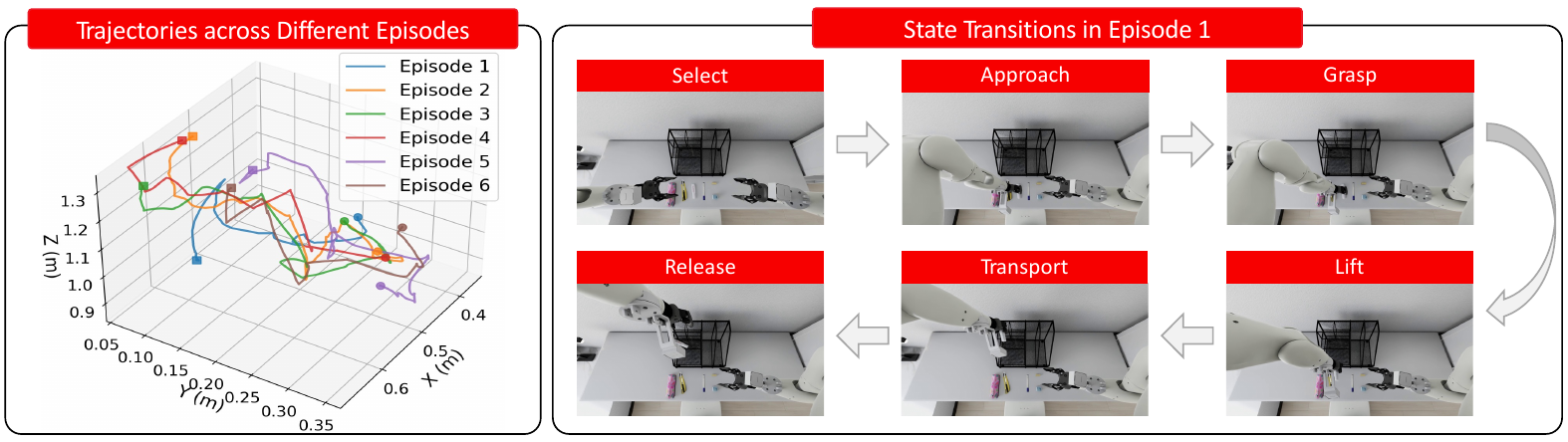}
    \caption{Synthetic Data Generation with an FSM-Based Method}
    \label{fig:fsm}
\end{subfigure}

\vspace{2mm}

\begin{subfigure}[t]{0.98\textwidth}
    \centering
    \includegraphics[width=\linewidth]{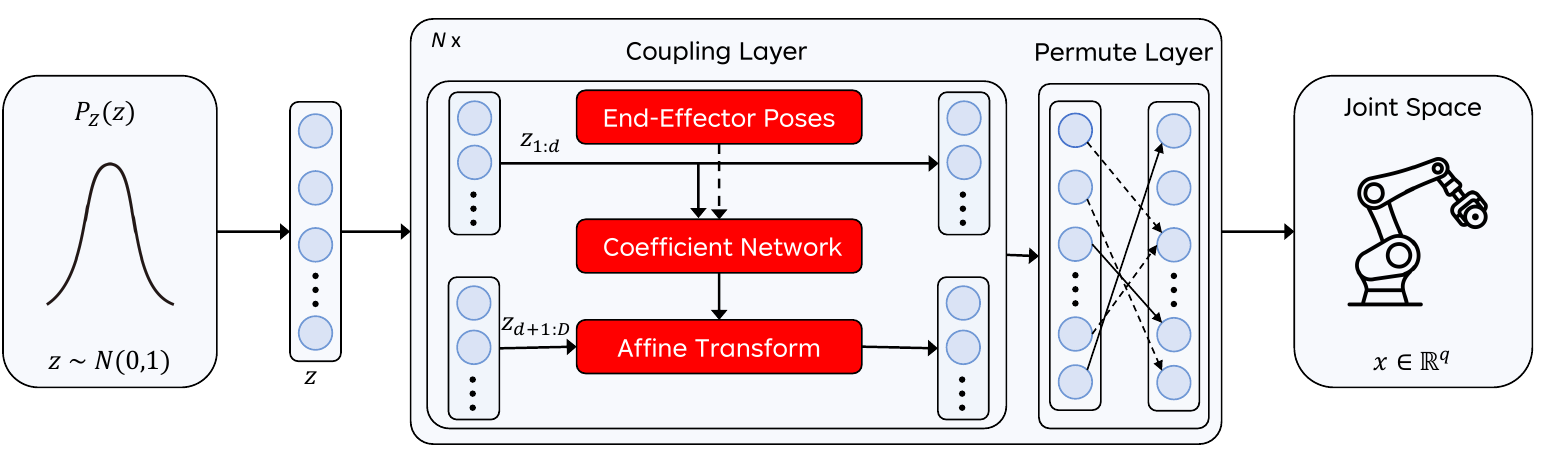}
    \caption{Synthetic Data Generation with an IKFlow-Based Method}
    \label{fig:ikflow}
\end{subfigure}

\vspace{2mm}

\begin{subfigure}[t]{0.98\textwidth}
    \centering
    \includegraphics[width=\linewidth]{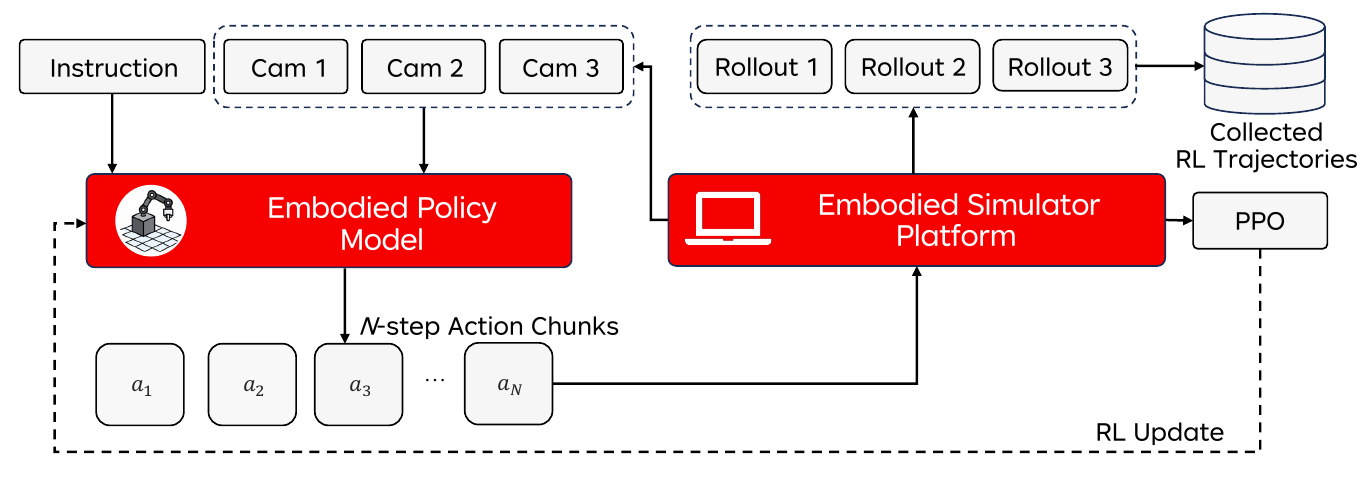}
    \caption{Synthetic Data Generation with an RL-Based Method}
    \label{fig:rl}
\end{subfigure}

\caption{
\textbf{Simulation-Enabled Synthetic Data Generation Methods.}
Four complementary pipelines are shown for generating robot trajectories in simulation: teleoperation, FSM-based generation, IKFlow-based augmentation, and RL-based autonomous data collection.
}
\label{fig:asset_alignment_reference}
\end{figure*}


\paragraph{Teleoperation.}
Teleoperation is a widely used paradigm for collecting robot demonstrations,
where human operators control robots in the simulator through interfaces such as a 6-DoF controller,
XR headsets,
and keyboards.
As illustrated in Figure~\ref{fig:tele},
operators can use dedicated buttons and knobs to control the gripper pose and
opening width,
enabling intuitive manipulation.
During operation,
human inputs are translated into robot control commands,
enabling the robot to perform manipulation tasks while recording observations,
robot states,
and action trajectories.
The collected trajectories can be further replayed,
processed,
and augmented to construct large-scale datasets for robot learning.
Teleoperation provides precise and natural demonstrations,
serving as high-quality seed trajectories for subsequent data augmentation and
policy learning in simulation.

\paragraph{FSM-based Method.}
For scenarios where demonstration trajectories are unavailable,
we adopt an FSM-based data synthesis method.
The approach first decomposes a manipulation task into a sequence of predefined
states based on task-specific prior knowledge,
such as object selection,
approach,
grasp,
transport,
and release.
Given object poses and environmental constraints,
target end-effector poses are generated for each state,
and an inverse kinematics solver is used to obtain feasible joint
configurations.
By sequentially executing these states,
complete manipulation trajectories can be automatically synthesized.
As illustrated in Figure~\ref{fig:fsm},
different episodes produce distinct trajectory distributions,
while each episode follows a fixed FSM structure.
Without requiring demonstration data,
this method can rapidly generate an initial task dataset for policy training and evaluation.

\paragraph{IKFlow-based Method.}
Once a small number of feasible demonstrations are available,
we employ an IKFlow-based augmentation strategy to expand the dataset,
as illustrated in Figure~\ref{fig:ikflow}.
IKFlow models the conditional distribution between end-effector poses and joint
configurations,
enabling diverse inverse-kinematics solutions to be generated efficiently for
the same task.
In practice,
a single successful demonstration serves as a seed trajectory,
from which multiple kinematically valid joint-space candidates are obtained
through latent-space sampling.
These candidate solutions are further connected through trajectory continuity
constraints and motion optimization to produce smooth and executable
trajectories.

\paragraph{RL-based Method.}
To further enhance data generation in complex environments,
we develop a simulation-based RL framework for autonomous trajectory
collection,
as illustrated in Figure~\ref{fig:rl}.
Existing demonstrations and simulation environments are first used to construct
training tasks,
while reward functions guide policy learning toward successful task completion.
The trained policy is then deployed under randomized scene configurations,
object properties,
and initial conditions to continuously generate diverse trajectories.
During this process,
both successful and failed trajectories are recorded together with their
corresponding failure modes and organized into a contrastive experience buffer.
This enables the policy to learn from informative positive and negative
examples,
thereby improving data diversity,
robustness,
and task generalization beyond what can be achieved with human demonstrations
alone.

\subsubsection{Consistency Calibration}
\label{sec:simulation-human-refinement}

\begin{figure}[htbp]
\centering
\includegraphics[width=\linewidth]{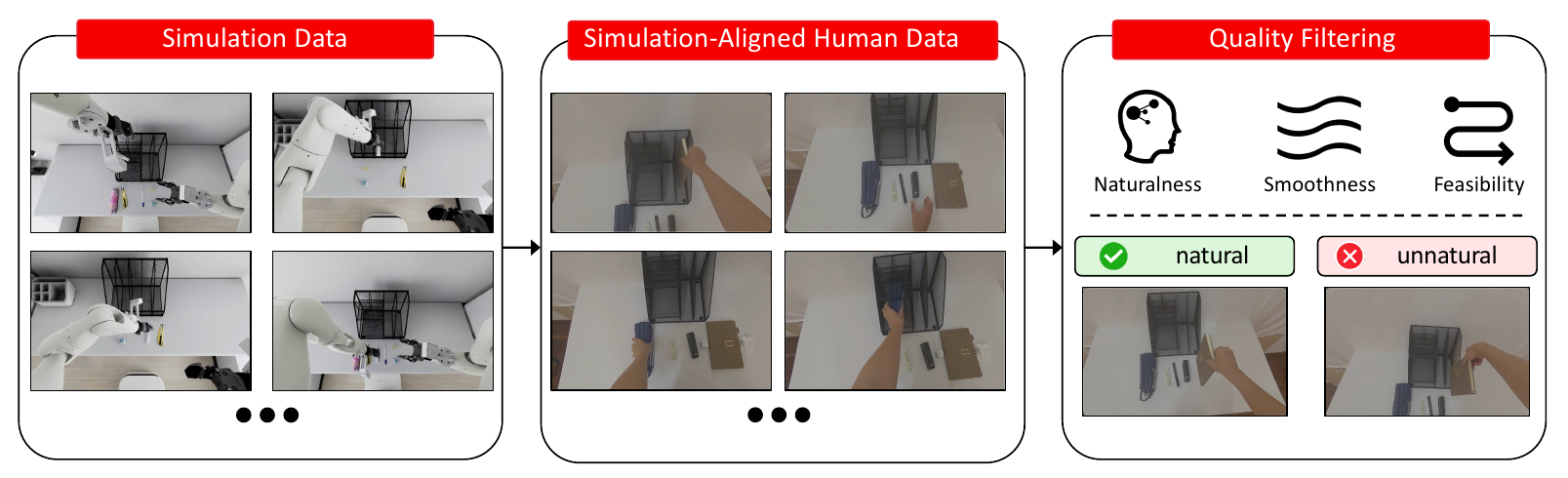}
\caption{
\textbf{Simulation $\rightarrow$ Human Trajectory Quality Refinement.}
Robot trajectories generated in simulation are projected into human-hand space for embodied inspection, enabling human operators to detect unnatural motion patterns and obtain aligned robot-human trajectory data from the same episode.
}
\label{fig:s2h2}
\end{figure}

Simulated robot trajectories often exhibit unnatural or even
physically implausible motion patterns due to the lack of human motion priors.
Although recent works such as
CRISP~\cite{lim2026robotsinnercriticselfrefinement} employ Vision-Language
Models (VLMs) to automatically evaluate the appropriateness and naturalness of
generated behaviors in simulation,
their judgments primarily rely on visual observations and semantic
understanding,
making it difficult to capture the rich motor intuition and experience humans
acquire through physical interaction.
Motivated by these observations,
we propose a Simulation $\rightarrow$ Human trajectory quality refinement
method,
as illustrated in Figure~\ref{fig:s2h2}.
The goal is to further improve the quality of trajectories generated in simulation by obtaining human data aligned with each simulation episode.
The core idea is to convert simulated robot trajectories into human-hand space and ask human operators to trace the mapped hand-space trajectories from an embodied perspective, thereby identifying unnatural or implausible motion patterns through direct embodied feedback.
Human operators can leverage rich physical intuition and motor experience when
simulating trajectory execution,
yielding fine-grained assessments of trajectory plausibility.
Specifically,
we first map the robot end-effector trajectories from the simulator to the human hand
workspace,
enabling human operators to track and reproduce trajectories from a
first-person perspective.
During this process, human operators can perceive awkwardness and implausible motion patterns in certain trajectories, which often reveal strategic deficiencies in simulated trajectories that violate human motor intuition.

By systematically collecting and analyzing embodied feedback from human operators on simulated trajectories,
we observe that the discrepancies between simulated trajectories and natural human
operation manifest as strategic deviations across multiple dimensions.
These include approach strategy deviation,
where simulated trajectories tend to select kinematically direct paths when
approaching target objects,
whereas human operators prefer approach directions that preserve greater
operational freedom for downstream actions.
They also include temporal strategy deviation,
where transitions between key operation phases in simulated trajectories often
lack the anticipatory adjustments that humans naturally perform,
resulting in insufficient motion fluency.
Other deviations in motion patterns may arise when simulation-based planning algorithms converge to locally optimal but globally suboptimal solutions.
Based on these naturalness criteria distilled from human embodied feedback,
we construct a trajectory naturalness assessment framework.
We then perform quality filtering on the simulated dataset,
removing trajectory samples that violate human motor intuition,
thereby obtaining a higher-quality training dataset.

The key advantage of the proposed method lies in establishing a closed-loop
pathway from simulation data to human embodied feedback and back to simulation data
quality.
Human operators' physical intuition provides criteria for
global-strategy naturalness assessment that neither per-frame evaluators nor
VLMs can replace.
The filtered dataset provides higher-quality candidate data for downstream manipulation-model training in both simulated and real-world settings.


\subsubsection{Robot-Human Data Alignment}
\label{sec:Robot_Human_Data_Alignment}

Beyond trajectory quality filtering, the Simulation $\rightarrow$ Human step
also provides a mechanism for constructing aligned robot-human data. Starting from a simulated robot episode, we can export the same underlying task execution in two complementary forms: a robot-centered trajectory that preserves the robot embodiment, control interface, and camera observations, and a human-form trajectory obtained by projecting the robot end-effector motion into human-hand space. As a result, each simulated episode yields a paired robot-human sample with shared task semantics, object states, temporal structure, and success labels.

This paired structure is valuable for representation alignment in the intermediate training stage before deployment-oriented fine-tuning, helping bridge human-centric priors and robot-executable actions.
Similar in spirit to human-robot data construction in EgoScale~\cite{zheng2026egoscale}, the aligned pairs
allow the model to observe how the same manipulation intent appears under
different embodiments and viewpoints. The robot trajectory provides deployment
grounding, while the human-form trajectory provides an embodiment that is closer
to large-scale human demonstration data. Training on such pairs encourages the
policy representation to effectively bridge human-centric manipulation priors and
robot-executable actions, instead of treating human and robot data as two
unrelated sources.

\section{Human $\rightarrow$ Simulation $\rightarrow$ Robot}
\label{sec:human_sim_robot}

Modern robot foundation models, including Vision-Language-Action (VLA)~\cite{brohan2023rt}, Vision-Action (VA)~\cite{brohan2022rt}, and World-Action Models (WAM)~\cite{ye2026worldaction}, require data that is both scalable and deployment-faithful.
Large robot datasets have improved generalist policies, but real-robot demonstrations remain expensive to collect across diverse scenes, embodiments, and tasks~\cite{brohan2022rt,oneill2024openx,khazatsky2024droid}.
Egocentric human videos offer a much larger and cheaper source of manipulation experience~\cite{grauman2022ego4d}, yet robots cannot directly execute them.
This motivates a simulation-mediated data pathway that converts scalable human demonstrations into physically feasible and robot-centered training data.

\begin{figure}[!tp]
\centering
\includegraphics[width=\linewidth]{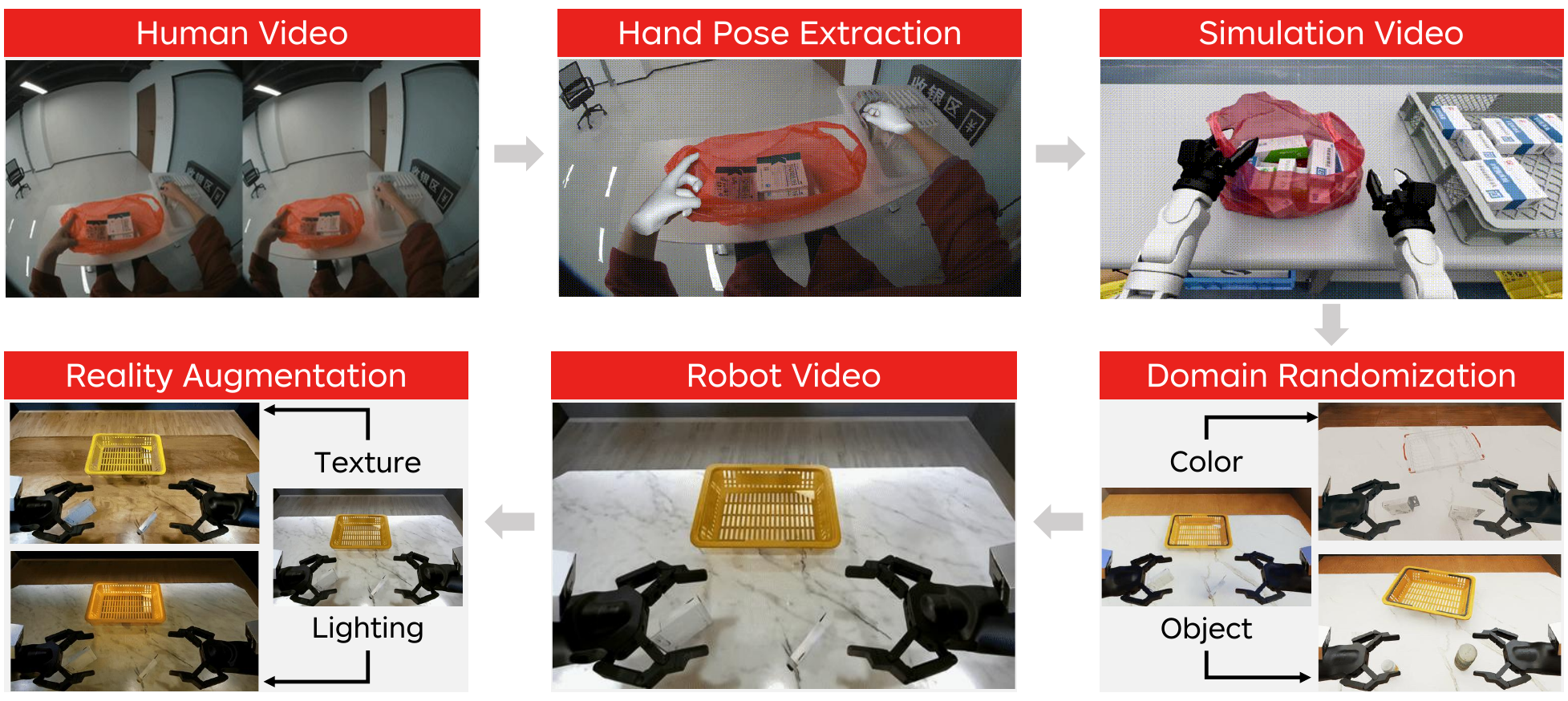}
\caption{\textbf{Human $\rightarrow$ Simulation $\rightarrow$ Robot Toolchain.}
\textbf{Stage~I: Human $\rightarrow$ Simulation.}
Egocentric human operation videos are processed to extract task-relevant hand pose cues and reconstruct the surrounding task scene, which are then converted into a sim-ready task representation.
\textbf{Stage~II: Simulation $\rightarrow$ Robot.}
The reconstructed task execution is converted into robot-centered manipulation replay, where simulated robot videos are expanded through domain randomization over object identity, color, and scene configuration, and further enhanced by reality augmentation to improve texture, illumination, and visual realism.}
\label{fig:h2s2r_pipeline}
\end{figure}

This section focuses on data generation, where the simulator bridges large-scale human videos and robot-centered data through a Human~$\rightarrow$~Simulation~$\rightarrow$~Robot toolchain.
Specifically, the toolchain converts human demonstrations into feasible robot trajectories and visually realistic robot videos.
Figure~\ref{fig:h2s2r_pipeline} summarizes the overall data pathway.
It starts from egocentric human videos, extracts task-relevant hand pose cues, and uses the simulator to obtain robot-centered trajectories and rendered videos.
The rendered results can then be expanded by domain randomization in the simulator and further adapted by reality augmentation to reduce the visual gap to real-robot deployment.
Simulation is necessary because direct human-to-robot retargeting is often unreliable.
Human hands and humanoid end-effectors differ in kinematics, contact geometry, and feasible force profiles, so naively transferred motions may violate joint limits, cause self-collisions, or produce physically inconsistent contacts.
We therefore first lift human demonstrations into metric 3D motion, execute them inside the simulator, and then export feasible robot trajectories together with robot-view rendered videos.
Domain randomization changes scene factors such as object instances, colors, and layouts, while reality augmentation improves the visual realism of the generated robot videos by adapting textures, illumination, and appearance.
In this way, the simulator acts both as a physical feasibility filter and as a data amplifier, while reality augmentation makes the resulting videos closer to the visual distribution of real-robot deployment.
The following subsections detail the Human $\rightarrow$ Simulation conversion in Sec.~\ref{sec:human_to_simulation} and the Simulation $\rightarrow$ Robot conversion in Sec.~\ref{sec:simulation_to_robot}.


\subsection{Human $\rightarrow$ Simulation}
\label{sec:human_to_simulation}

The Human $\rightarrow$ Simulation stage serves as the front-end component of the Human $\rightarrow$ Simulation $\rightarrow$ Robot pipeline.
Its goal is not to directly copy human hand motion onto the robot, but to extract task-relevant motion cues from egocentric human videos and place them into the simulator context, where robot feasibility can be checked.
In the current system, this stage is organized as a conversion workflow that focuses on hand motion, object-level task cues, and an editable simulator setup.
Uncertain cases are manually corrected or verified before being used for robot rollout generation.

\paragraph{Human motion and task information.}
The input is an egocentric manipulation video, optionally accompanied by depth observations, camera calibration, object annotations, or selected key frames.
From the RGB stream, we use HaMeR~\cite{pavlakos2024hamer} to estimate the demonstrator's hand pose.
These estimates are used as motion priors rather than direct robot actions.
Instead of preserving the full human hand articulation, we extract a task-space description, including wrist motion, approach direction, grasp or release phase, and candidate hand--object interaction regions.
This abstraction reduces the dependence on human-specific kinematics and keeps the information most relevant to robot manipulation.
The task context is then instantiated in the simulator.
The manipulated object, target container or support region, camera configuration, and robot embodiment are specified or selected from available assets.
Object poses and task semantics are aligned with the demonstration when reliable estimates are available.
When some quantities cannot be confidently estimated from the video alone, they are treated as editable parameters rather than fixed outputs of the perception system during rollout construction and validation.

\paragraph{Constraint-aware robot rollout.}
After the motion prior and task context are available, the demonstration is instantiated in the simulator and adapted to the target robot.
The human trajectory provides a coarse task prior, while the simulator checks whether a corresponding robot rollout is feasible under the robot embodiment.
This follows the motivation of recent human-video-to-robot-skill pipelines, where the simulator is used to bridge the embodiment gap between human demonstrations and robot manipulation~\cite{hsieh2025dexman}.
The rollout is filtered by kinematic, safety, and task-level constraints.
These constraints cover joint limits, reachable workspace, gripper range, collision avoidance, object accessibility, grasp or support stability, and target-region completion.
If a rollout violates these constraints, it can be locally adjusted,
manually reviewed, or excluded from the executable training set.
This prevents visually plausible but physically invalid human motions from being used as robot demonstrations.

\paragraph{Front-end products for render-and-overlay.}
The steps above feed the full-simulation route, where the robot is reconstructed and rendered inside the simulator.
The same stage can also prepare the inputs for the render-and-overlay route, which later composites the robot onto the original video in Sec.~\ref{sec:simulation_to_robot}.
For that route, two front-end products are produced here.
First, the demonstrator's hand is removed from the egocentric video with a diffusion-based video inpainting model, which fills the occluded region and yields a clean, hand-free background that preserves the original scene appearance.
Second, the reconstructed 3D hand pose annotations are used to solve the target robot's arm degrees of freedom through inverse kinematics (IK), so that the robot end-effector follows the demonstrated wrist motion while respecting the robot embodiment and the same feasibility constraints above.

\paragraph{Outputs.}
For each accepted demonstration, the Human $\rightarrow$ Simulation stage produces a robot-centered simulated episode.
The episode contains robot states, end-effector poses, gripper commands, camera observations, object poses, contact events, and task annotations.
Compared with the original human video, the resulting data is expressed in the robot's action and observation space.
Compared with a purely visual reconstruction, it is tied to an executable robot state and can be replayed, perturbed, and rendered under controlled conditions.
These robot-centered rollouts serve as seed data for the following Simulation $\rightarrow$ Robot stage, where domain randomization and reality augmentation are used to produce more diverse and realistic robot-view videos for downstream training.

\subsection{Simulation $\rightarrow$ Robot}
\label{sec:simulation_to_robot}

The Simulation $\rightarrow$ Robot stage converts the robot-centered simulated episodes from Sec.~\ref{sec:human_to_simulation} into visually diverse robot-view videos for downstream training.
As illustrated in Figure~\ref{fig:h2s2r_pipeline}, this stage starts from videos rendered in the simulator and produces two types of outputs: domain-randomized videos and realism-augmented robot-view videos.
We implement this stage through domain-randomized rendering, reality augmentation based on Cosmos Transfer~\cite{ali2025world}, and appearance-level visual generalization.

\paragraph{Domain-randomized simulation rendering.}
We first expand each simulated episode by randomizing task-relevant and visual factors inside the simulator.
At the task level, object poses, container locations, and object combinations are varied while maintaining reachability and task feasibility.
At the visual level, object colors, container colors, distractor objects, and local textures are randomized.
For example, in the basket manipulation task shown in Figure~\ref{fig:h2s2r_pipeline}, changing the basket color or the manipulated object produces multiple valid videos from the same task structure.
This follows the standard domain-randomization principle: increasing the coverage of the simulation distribution can reduce the visual and state mismatch encountered during real deployment~\cite{tobin2017domain,peng2018sim2real}.
The output is a set of domain-randomized videos with controlled variations in object identity, color, pose, and local scene configuration.
Because these videos are generated inside the simulator, they remain aligned with robot states, object poses, camera parameters, and task labels, making them usable as paired trajectory-video data for robot foundation model training.

\paragraph{Reality augmentation with Cosmos Transfer.}
Domain randomization improves diversity, but simulated videos may still differ from those captured by real robot cameras.
We therefore use Cosmos Transfer~\cite{ali2025world} as a reality-augmentation module to translate simulated clips into more realistic robot-view videos.
In our pipeline, the RGB video provides the motion and scene layout, while structural signals such as depth, edges, or segmentation can be used to preserve the spatial structure of the scene during visual transfer.
This step adapts the visual appearance of the video while preserving the task content, including robot motion and object movement. Quality checks are applied to ensure that the transferred videos remain consistent with the original simulation labels.
Thus, the same executable trajectory can provide both physically grounded labels and visually enhanced observations for downstream model training.

\paragraph{Appearance-level visual generalization.}
Beyond one-to-one reality augmentation, the same transfer module can generate plausible appearance variants from a fixed rollout, such as changes in tabletop material, illumination, and global light tone.
This complements domain randomization in simulation: the simulator varies physically explicit factors such as object pose, object identity, and container color, while Cosmos-based transfer adjusts higher-level appearance factors that are costly to author manually.

\paragraph{Render-and-overlay alternative.}
The render-and-overlay route completes the alternative path by turning the two front-end products from Sec.~\ref{sec:human_to_simulation}, namely the hand-free inpainted video and the IK-solved robot arm trajectory, into a robot-view video without a fully simulated scene.
The IK-solved robot arm is rendered and overlaid onto the inpainted video, aligning it with the original scene geometry and camera viewpoint.
A video editing model such as Cosmos Transfer~\cite{ali2025world} then corrects occlusion boundaries, contact shadows, and other physical details, harmonizing the overlaid robot with the surrounding scene.
Because the background is the real video rather than a rendered scene, this route preserves the original appearance and illumination, while still producing robot-centered videos paired with IK-solved trajectories.

The Simulation $\rightarrow$ Robot stage therefore produces domain-randomized videos and reality-augmented robot-view videos.
Together, they enhance controllable diversity and visual realism while preserving the physical structure and privileged annotations provided by the underlying digital twin.



\section{Scalable Simulation Infrastructure}
\label{sec:scalable_simulation_infrastructure}

JoyAI-Sim requires a unified infrastructure that can scale simulation execution, scene construction, and task development. This section introduces the cloud-native execution layer and the NVIDIA Isaac Lab Arena~\cite{isaaclab-arena2025} framework used to expand validated simulation prototypes into a diverse task suite.

\subsection{Cloud-Native Infrastructure for Data Generation and Evaluation}
\label{sec:cloud_deployment}

\begin{figure}[!htbp]
\centering
\includegraphics[width=\linewidth]{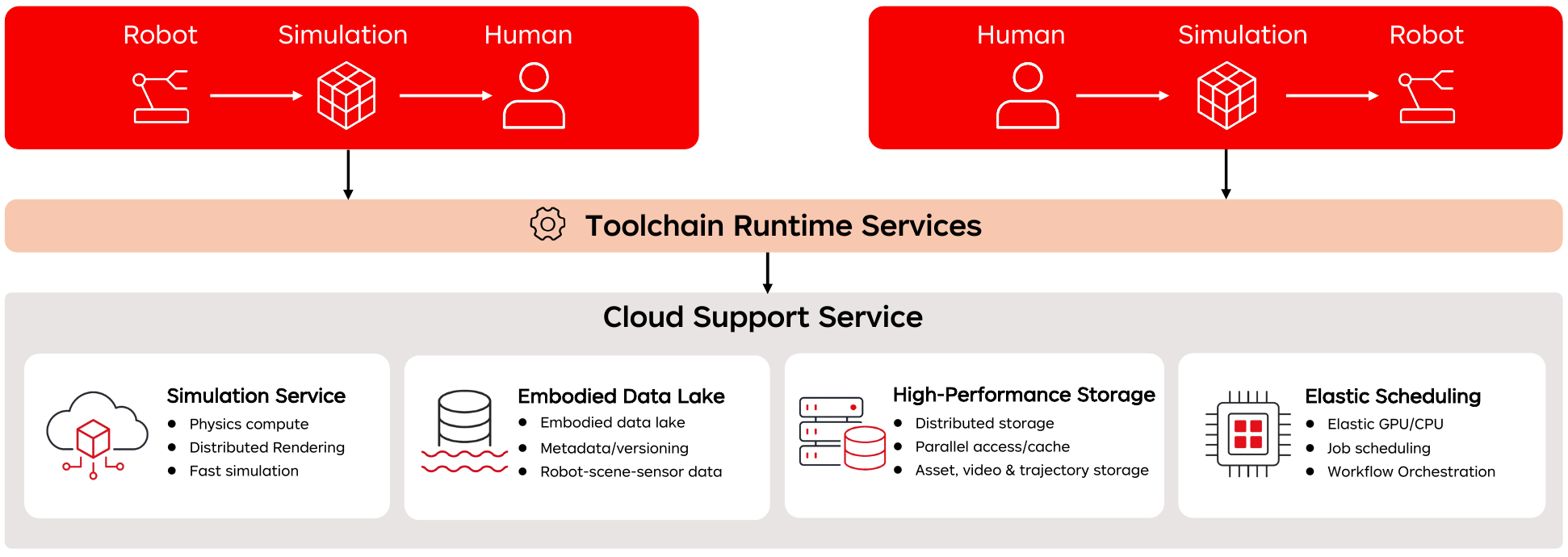}
\caption{
\textbf{Cloud-Native Infrastructure for Embodied Data Generation and Evaluation.}
JD Cloud provides the cloud-side execution substrate for the Robot $\rightleftharpoons$ Simulation $\rightleftharpoons$ Human workflow.
The infrastructure integrates simulation service, embodied data lake, high-performance storage, and elastic scheduling.
}
\label{fig:jd_cloud_infra}
\end{figure}

JoyAI-Sim organizes simulator rollout, data conversion, robot-view rendering, reality augmentation, data management, and policy screening as a cloud-native execution layer on JD Cloud\footnote{\url{https://www.jdcloud.com/}}.
The cloud deployment is supported by JoyBuilder 2.0 service entries\footnote{\url{https://docs.jdcloud.com/cn/jdaip/product-overview}}, including the embodied simulation service for cloud-side simulation execution\footnote{\url{https://docs.jdcloud.com/cn/jdaip/create-embodiedsimulation}} and the service for reality augmentation\footnote{\url{https://docs.jdcloud.com/cn/jdaip/Notebook-Cosmos-Transfer}}.
As illustrated in Figure~\ref{fig:jd_cloud_infra}, this layer sits below the bidirectional Robot $\rightleftharpoons$ Simulation $\rightleftharpoons$ Human workflow and provides shared toolchain runtime services for both data generation and evaluation.
Its purpose is not to introduce an additional operational stage, but to provide a common execution substrate through which the two pathways can be scaled and managed under consistent runtime configurations.

Corresponding to the four cloud support services in Figure~\ref{fig:jd_cloud_infra}, the cloud-side infrastructure consists of four functional components.
The simulation service provides simulation and rendering engine support, including parallel physics computation, distributed rendering, and high-throughput simulation execution.
It is responsible for scene preparation, robot embodiment loading, camera configuration, task rollout execution, and robot-view video rendering.
The elastic scheduling and governance component provides GPU/CPU elastic resources, distributed job scheduling, and workflow orchestration, allowing rollout, rendering, and augmentation jobs to be executed in parallel under controlled runtime versions.
The high-performance storage component supports parallel access and caching for large-scale assets, videos, and trajectory files.
The embodied data lake unifies robot, scene, sensor, and trajectory data with metadata indexing and dataset versioning, so that generated data and evaluation results remain traceable across experiments.

This shared runtime is designed to support three capabilities for data generation and evaluation.
First, it provides an execution interface for batching simulation rollouts, rendering robot-view trajectories, and applying domain randomization over object instances, object poses, scene layouts, illumination, textures, backgrounds, and robot states.
Each accepted rollout can be exported with synchronized states, actions, object poses, camera observations, and task annotations, allowing a validated task setup to produce multiple physically grounded training examples rather than a single deterministic replay.
Second, it provides a standardized interface between simulation rendering and reality augmentation.
The simulator preserves physically explicit trajectories, privileged states, and structured annotations, while the realism-augmentation module is intended to reduce the visual gap to robot-camera observations without changing the underlying task execution.
Third, it provides a common configuration layer for policy screening by fixing runtime versions, assets, embodiments, camera models, success predicates, and output schemas across policy checkpoints.
Together, these components are intended to provide an extensible execution foundation for scalable data generation, visual enhancement, and controlled simulation-based evaluation across different deployment workloads.

\subsection{Isaac Lab Arena for Scalable Scene and Task Expansion}
\label{sec:isaac_lab_arena}

\begin{figure}[!htbp]
\centering
\includegraphics[width=\linewidth]{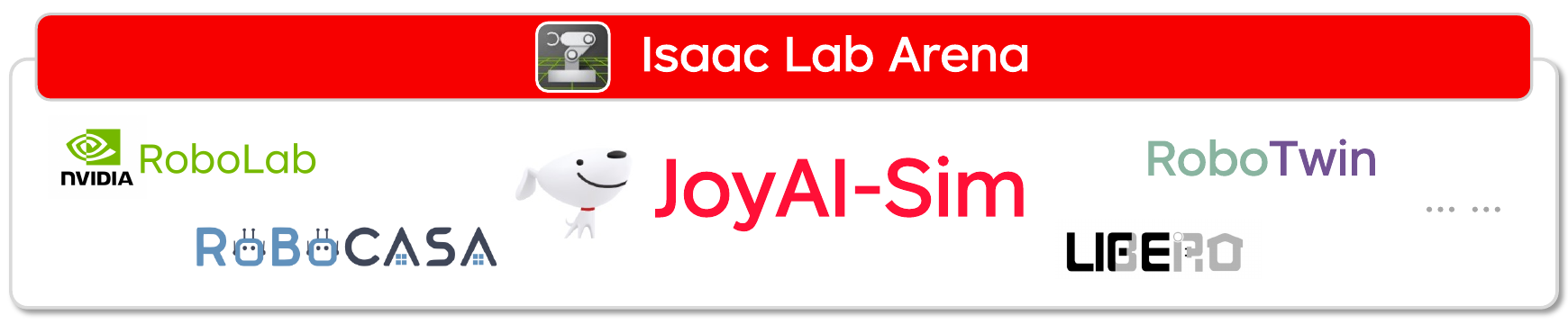}
\caption{
\textbf{JoyAI-Sim Integrated Within Isaac Lab Arena for Unified Embodied Simulation and Evaluation.}
Isaac Lab Arena provides a common interface for simulation environments and embodied AI benchmarks.
Within the broader Arena ecosystem, JoyAI-Sim complements simulation environments and embodied AI benchmarks such as RoboLab, RoboCasa, RoboTwin, and LIBERO, supporting standardized policy deployment and evaluation through a shared framework.
}
\label{fig:isaac_lab_arena}
\end{figure}

JoyAI-Sim integrates with the NVIDIA Isaac Lab Arena~\cite{isaaclab-arena2025} by adopting its compositional Scene--Embodiment--Task abstractions for environment construction and policy evaluation.
In this integration, a scene is assembled from independently registered assets, including manipulable objects, articulated objects, backgrounds, and other environmental elements.
An embodiment encapsulates the robot's physical configuration, sensors, observations, and action interfaces, while a task defines the objective, initialization and reset logic, event handling, termination and success conditions, and evaluation metrics.
JoyAI-Sim registers reconstructed digital-twin scenes and robot configurations through these interfaces, preserving the physical assets, spatial layouts, and task-relevant properties of the corresponding real-world evaluation settings.
Task definitions, observation and action spaces, reset procedures, success predicates, and evaluation metrics are then organized through the same Arena abstractions.
Figure~\ref{fig:isaac_lab_arena} illustrates how JoyAI-Sim is organized within
the broader Isaac Lab Arena ecosystem. This integration turns each reconstructed digital twin into a reusable environment specification and provides a consistent interface for extending its scenes and tasks.

Within this framework, JoyAI-Sim is designed to expand each validated scene--task prototype from a single configuration into a family of variants.
The planned expansion covers object instances and layouts, scene appearance and geometry, initial states, robot embodiments, and task parameters, allowing one manually validated setup to seed $N$ systematically generated environments.
This one-to-$N$ design supports multiple manipulation tasks with consistent
interfaces and evaluation criteria.
Together with the cloud-native execution layer, Arena enables scalable rollout
generation and reproducible policy evaluation.

\section{Experiments}
\label{sec:experiments}

We evaluate JoyAI-Sim from two complementary perspectives: model evaluation
and data synthesis.
From the model-evaluation perspective, we assess whether digital twins
reconstructed from real-robot settings preserve deployment-relevant evaluation
signals while enabling efficient and repeatable policy evaluation.
From the data-synthesis perspective, we examine whether simulation-generated
trajectories improve policy performance when combined with real-robot
demonstrations, and whether embodied human feedback further improves the
synthesized data by identifying and filtering unnatural manipulation strategies.
Together, these experiments characterize JoyAI-Sim as a simulation-centered
framework for reliable model evaluation and high-quality data synthesis.

\subsection{Digital-Twin Evaluation: Consistency, Efficiency, and Stability}
\label{sec:supp_sim_real_env_consistency}


\paragraph{Real-Simulation Consistency.} We assess real-simulation consistency at two complementary levels:
\begin{wrapfigure}[22]{r}{0.39\textwidth}
    \vspace{-10pt}
    \centering
    \includegraphics[width=\linewidth]{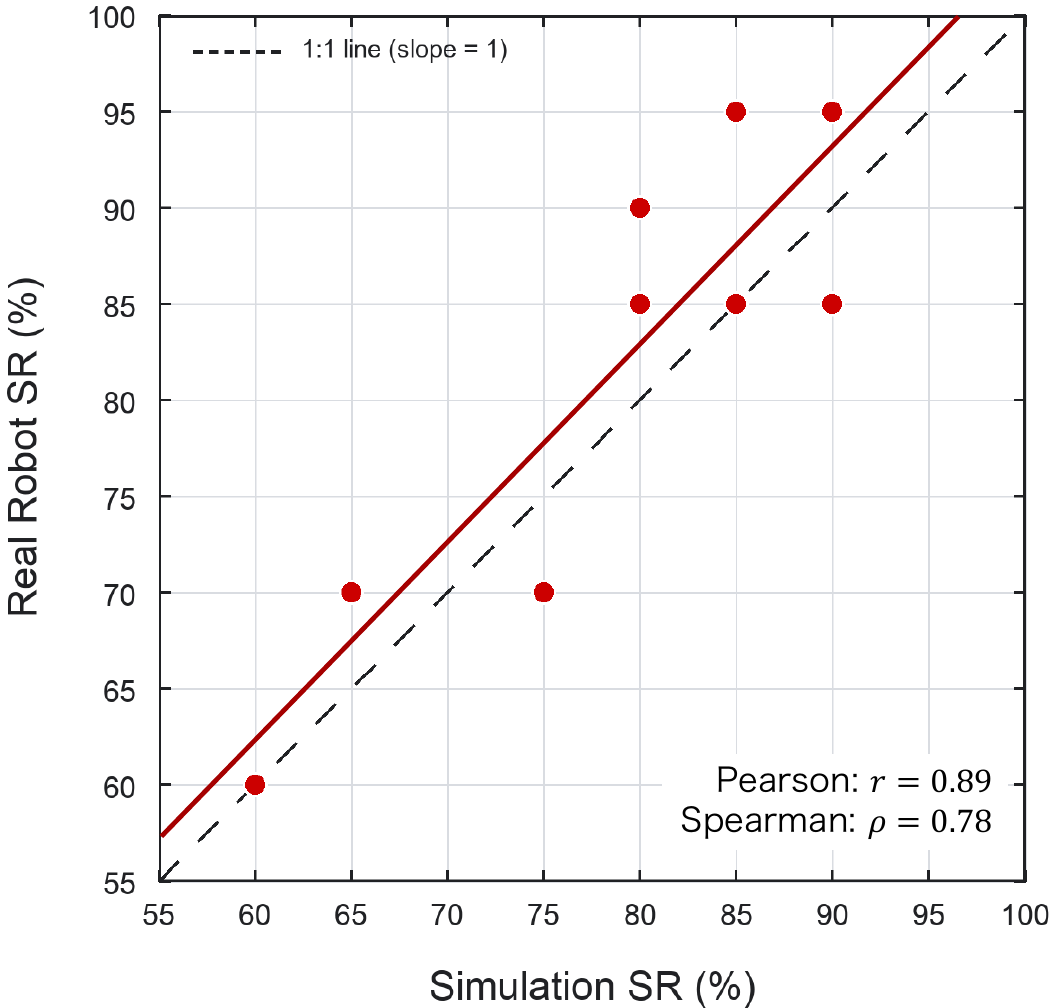}
    \vspace{-20pt}
    \caption{
    \textbf{Correlation Between Simulation and Real-Robot Success Rates.}
    Each point denotes one paired evaluation. The dashed black line is the $1{:}1$ reference line, while the solid red line is the least-squares linear regression fit. The correlation coefficient indicates strong sim-to-real consistency.
    }
    \label{fig:sim_real_corr}
\end{wrapfigure}
environment-level outcome reproduction and policy-level evaluation agreement.
At the environment level, we replay in simulation 100 matched real-robot
trajectories collected from the desk-tidying task, using the same task
definitions and object-level success criteria.
Following the scene and interaction alignment protocols described in
Sec.~3.1.2 and Sec.~3.1.3, each real-robot setup is reconstructed as a digital
twin by aligning the task-relevant scene layout, robot embodiment, control
interface, and episode lifecycle.
We replay the recorded control streams in simulation and compare the resulting
object placements with those observed during the corresponding real-robot
executions.
A replay is considered consistent when the task-relevant objects are placed in
the same target locations as in the real scene after task completion.
Among the evaluated trajectories, approximately 90\% of the simulation replays
reproduce object-placement outcomes consistent with their real-robot
counterparts.

At the policy-evaluation level, we evaluate checkpoints from different training
configurations and training steps in both the digital twin and on the real
robot, yielding paired simulation and real-robot success rates.
As shown in Figure~\ref{fig:sim_real_corr}, the two metrics exhibit a strong
positive linear association, with Pearson's $r=0.89$, while their rank-based
association is also positive, with Spearman's $\rho=0.78$.
The results indicate that checkpoints performing better in simulation generally
also perform better on the real robot.
Taken together, the trajectory-level agreement and policy-level correlation
show that the reconstructed digital twin reproduces task-relevant real-world
outcomes while preserving useful relative performance signals across candidate
policies.
This supports its use as a simulation-first screening environment before
selective real-robot evaluation.

\paragraph{Evaluation Efficiency and Stability.}
We evaluate the operational efficiency and stability of Simulation using two
metrics: evaluation throughput and repeated-run stability.
Throughput is measured under representative deployment configurations for each
evaluation platform, reflecting their achievable system-level capacity rather
than a hardware-normalized comparison.
The simulation platform uses 8 GPUs, with 8 parallel instances per GPU, yielding
64 concurrent environments.
Under this configuration, Simulation completes 128 evaluation episodes per hour,
compared with 40 episodes per hour using the physical-robot evaluation setup,
providing a 3.2$\times$ increase in overall evaluation throughput.
To assess stability, we conduct multiple independent evaluation runs and report
the standard deviation of the resulting success-rate estimates.
Simulation achieves a standard deviation of 3.16\%, compared with 5.07\% on the
physical robot, indicating more repeatable evaluation results under matched task
definitions.
Together, these results demonstrate that Simulation provides a high-throughput
and stable screening platform for efficiently evaluating candidate policies
before selective real-robot deployment.
Table~\ref{tab:supp_eval_efficiency_stability} summarizes the evaluation
throughput and stability results.

\begin{table}[t]
\centering
\small
\caption{\textbf{Efficiency and Stability of Simulation and Real-Robot Evaluation.}
Simulation provides faster and more repeatable evaluation while preserving the real-robot deployment signal.}
\label{tab:supp_eval_efficiency_stability}
\setlength{\tabcolsep}{5pt}
\renewcommand{\arraystretch}{1.08}
\begin{tabular}{lccc}
\toprule
\textbf{Metric}
& \textbf{Simulation}
& \textbf{Real Robot}
& \textbf{Simulation Improvement} \\
\midrule
Evaluation Throughput (episodes/hour) $\uparrow$
& \textbf{128}
& 40
& \textbf{3.2$\times$ higher} \\

Success-Rate Std. Across Runs (\%) $\downarrow$
& \textbf{3.16}
& 5.07
& \textbf{37.7\% lower} \\
\bottomrule
\end{tabular}
\end{table}

\subsection{Policy Gains from Simulation-Generated Data and Human Filtering}
\label{sec:exp_data_loop}

We investigate how simulation-generated data and human filtering improve
JoyAI-RA~\cite{zhang2026joyaira01foundationmodel} in downstream robot
manipulation tasks.
The evaluation uses tabletop organization as a representative manipulation
benchmark involving perception, grasping, object transport, and precise target
placement.
This multi-stage structure provides a practical setting for assessing whether
synthetic trajectories can complement real-robot demonstrations and support
effective physical execution.

\begin{figure*}[!htbp]
  \centering
  \includegraphics[width=\linewidth]{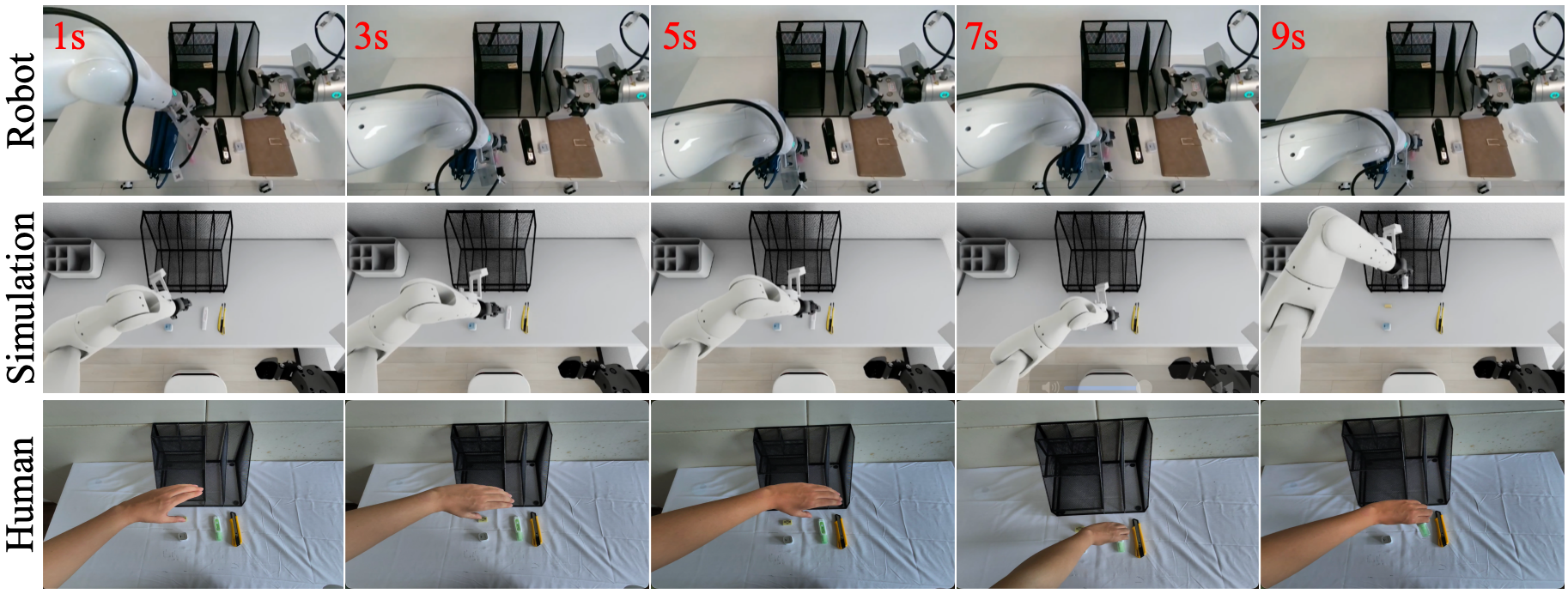}
  \caption{
  \textbf{Qualitative Visualization of Simulation $\rightarrow$ Human Trajectory Conversion.}
  Simulation-generated robot trajectories are converted into human-hand
  visualizations, allowing operators to assess motion naturalness and
  strategy quality from an embodied perspective.
  }
  \label{fig:simulation_to_human}
\end{figure*}

SimGen constructs synthetic training data through three complementary
mechanisms: teleoperation in simulation, automatic trajectory generation using
finite-state machines (FSMs), and IKFlow-based trajectory augmentation.
To further improve data quality, human filtering (HF) projects the resulting
robot end-effector trajectories into human-hand space, enabling operators to
inspect each execution from an embodied perspective.
As illustrated in Figure~\ref{fig:simulation_to_human}, this conversion makes
global strategy-level artifacts more apparent, including overly direct approach
paths, missing pre-adjustment motions, and abrupt transitions between
manipulation phases.
Trajectories exhibiting such unnatural behaviors are removed before policy
training.

We compare four training-data configurations:
\textbf{Real}, using only real-robot demonstrations;
\textbf{Real + SimGen}, augmenting the real data with trajectories synthesized
through the SimGen pipeline;
\textbf{Real + SimGen + HF}, retaining simulation-generated trajectories after
human filtering; and
\textbf{Real + SimGen + RL}, incorporating additional trajectories mined through
reinforcement learning.
These configurations isolate the contributions of simulation-generated data,
human filtering, and reinforcement-learning-based data mining.

\begin{figure}[!htbp]
\centering
\includegraphics[width=\linewidth]{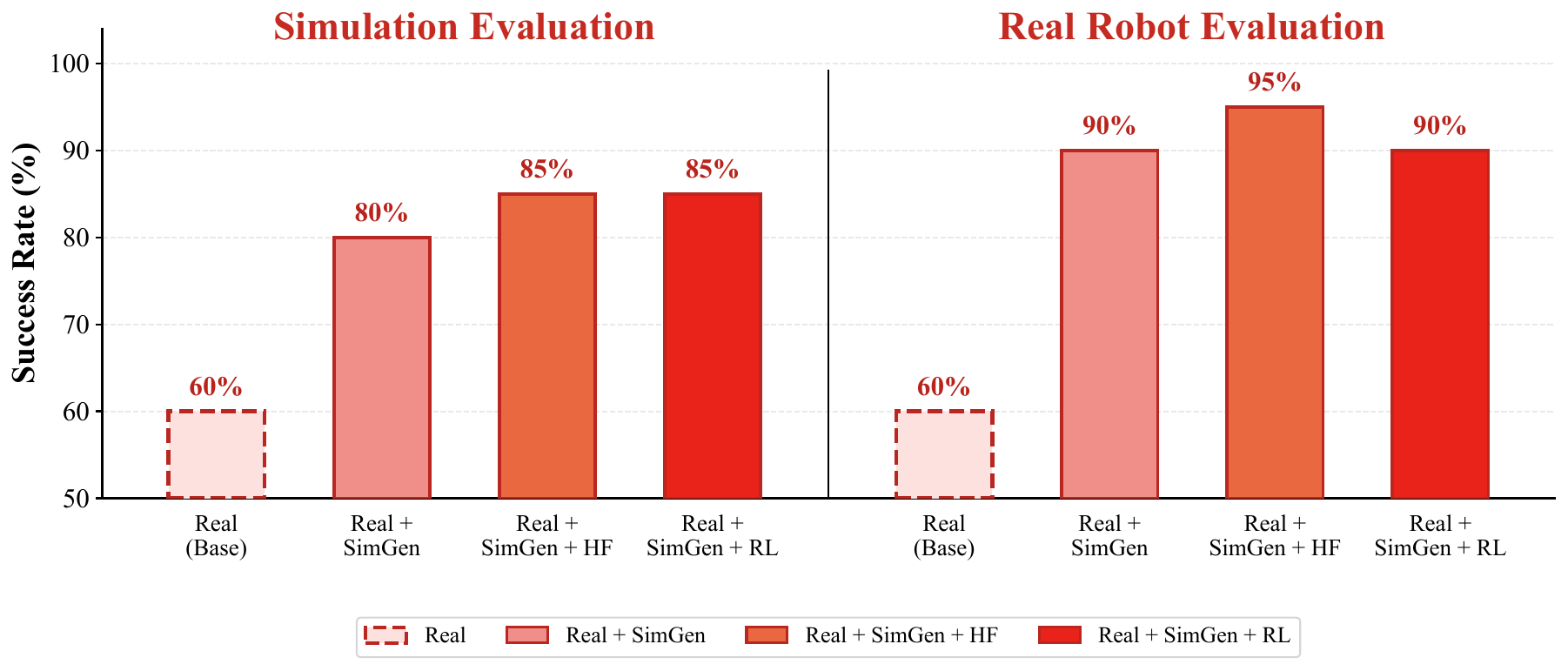}
\caption{
\textbf{Comparison of Policy Success Rates Under Different Training Strategies in Simulation and Real-Robot Experiments.}
Simulation-generated data improves policy performance, while human filtering
achieves the highest success rates in both evaluation settings.
}
\label{fig:main_results}
\end{figure}

Figure~\ref{fig:main_results} reports policy success rates in the matched
simulation and real-robot environments.
The Real-only baseline achieves a success rate of 60\% in both settings.
Adding SimGen increases the success rate to 80\% in simulation and 90\% on the
real robot, corresponding to absolute gains of 20 and 30 percentage points,
respectively.
Human filtering further raises the success rates to 85\% and 95\%, achieving the
best overall performance and improving upon SimGen alone by 5 percentage points
in both settings.

The reinforcement-learning variant reaches an 85\% success rate in simulation
and 90\% on the real robot.
Although it matches the simulation performance of human filtering, it provides
no additional gain over SimGen alone in the real-robot evaluation.
Overall, the results identify SimGen as the primary source of policy improvement,
while the consistent gains from human filtering demonstrate the value of
removing trajectories with unnatural embodied strategies.
Within this benchmark, the strong real-robot performance further indicates that
simulation-generated experience can effectively support physical deployment.

\section{Conclusion}
\label{sec:conclusion}
JoyAI-Sim is a simulation-enabled interconversion
toolchain for the embodied data pyramid. JoyAI-Sim formulates the Robot $\rightleftharpoons$ Simulation $\rightleftharpoons$ Human paradigm, using simulation as the central alignment layer between robot data and human data.
This bidirectional formulation is instantiated through two complementary
pathways. In the Robot $\rightarrow$ Simulation $\rightarrow$ Human pathway,
real-robot tasks and success criteria anchor calibrated digital-twin evaluators,
in which simulation enables scalable policy evaluation, while synthesized trajectories can be inspected with human embodied feedback before use as candidate training data.
In the Human $\rightarrow$
Simulation $\rightarrow$ Robot pathway, egocentric human demonstrations are
lifted into the simulator, checked under robot physical constraints, and
converted into robot-centered trajectories and robot-view observations for
downstream policy training.
Together, these pathways form a simulation-centered data loop that reduces the
cost and variance of repeated evaluation while supporting scalable robot data
generation.
The reconstruction, simulation, rendering, and realism-enhancement modules are
organized as services on JD Cloud, turning the toolchain into reusable
infrastructure for robot data production and deployment-oriented evaluation.
JoyAI-Sim further integrates with NVIDIA Isaac Lab Arena to organize
reconstructed environments, robot embodiments, and task definitions under a
unified interface, supporting reusable benchmark construction and scalable task development.

\clearpage
\begin{appendices}
\section{Additional Real-Robot Generalization Evaluation}
\label{app:real-robot-generalization-observations}

\begin{figure}[!htbp]
    \centering
    \begin{subfigure}{\linewidth}
        \centering
        \includegraphics[width=0.85\linewidth,height=0.36\textheight,keepaspectratio]{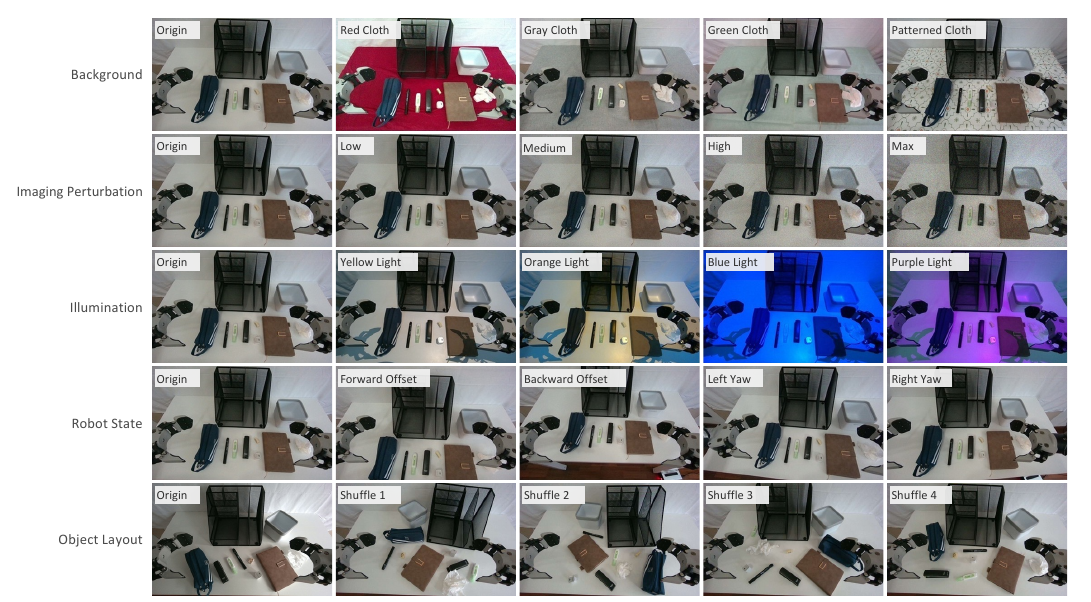}
        \caption{Generalization Axes in Study Room}
        \label{fig:real_robot_generalization_study}
    \end{subfigure}
    \vspace{0.15em}
    \begin{subfigure}{\linewidth}
        \centering
        \includegraphics[width=0.85\linewidth,height=0.36\textheight,keepaspectratio]{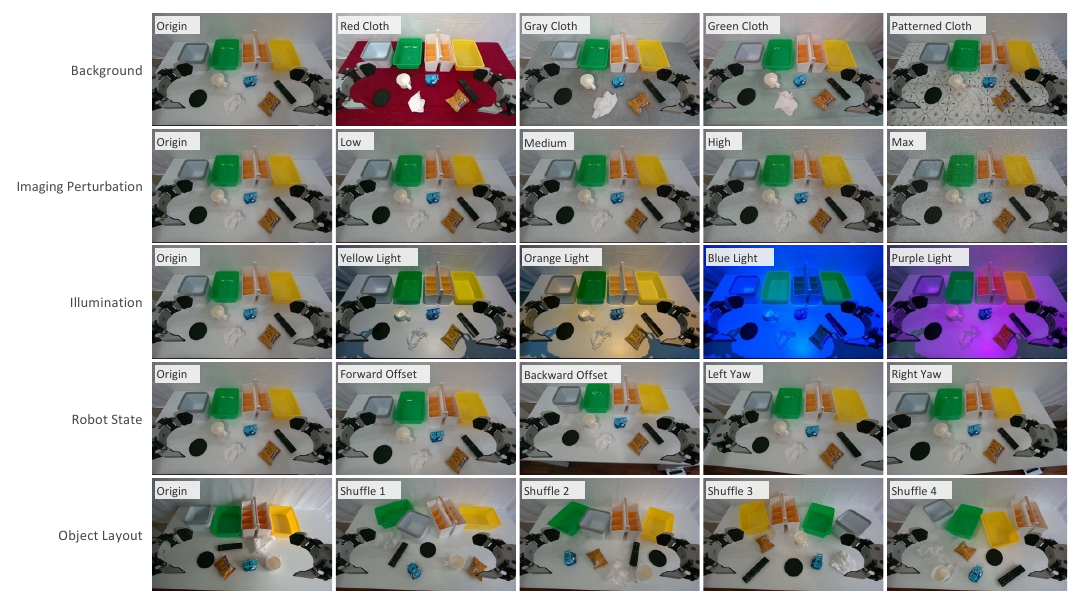}
        \caption{Generalization Axes in Living Room}
        \label{fig:real_robot_generalization_living}
    \end{subfigure}
    \caption{
    \textbf{Representative Real-World Observations for Generalization Evaluation.}
    Real-robot observations are captured under controlled real-world generalization settings for study-room and living-room tidy-up tasks.
    }
    \label{fig:real_robot_generalization}
\end{figure}

Figure~\ref{fig:real_robot_generalization} provides representative head-camera observations from the physical robot for the controlled real-world generalization settings discussed in Section~\ref{sec:JoyAI-Sim_real}.
These observations span tidy-up tasks in study rooms and living rooms with diverse visual variations.

\section{Additional Sim-Ready Asset Statistics}
\label{app:sim-ready-assets}

\begin{figure}[!htbp]
    \centering
    \includegraphics[width=1.0\linewidth]{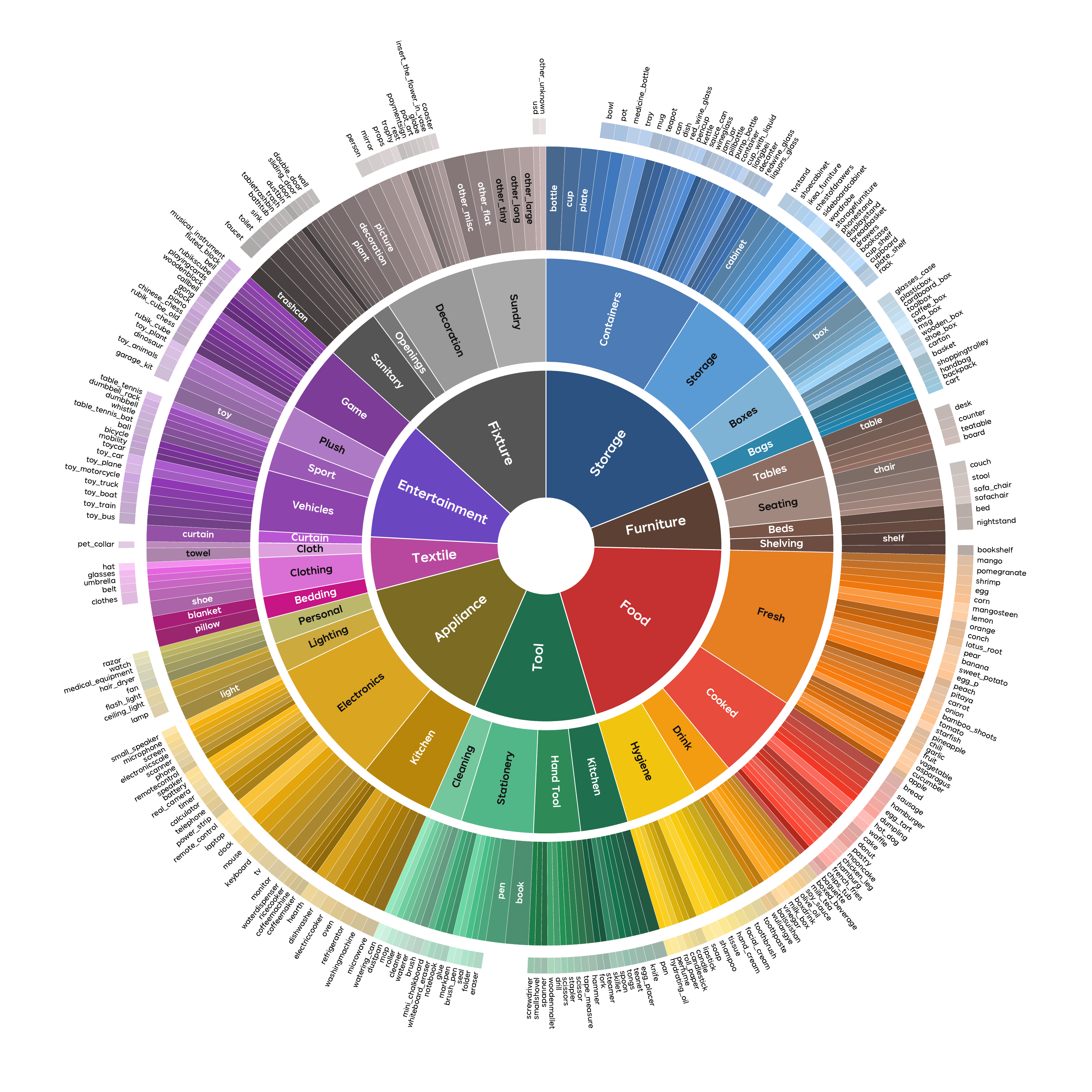}
    \caption{\textbf{Hierarchical Distribution of Sim-Ready Assets.}
The sunburst chart organizes the asset library by scene-level category, functional role, and fine-grained object class, providing an overview of the semantic coverage of the simulator asset space.}
    \label{fig:app_sunburst_merged}
\end{figure}

\begin{figure}[!htbp]
    \centering
    \includegraphics[width=1.0\linewidth]{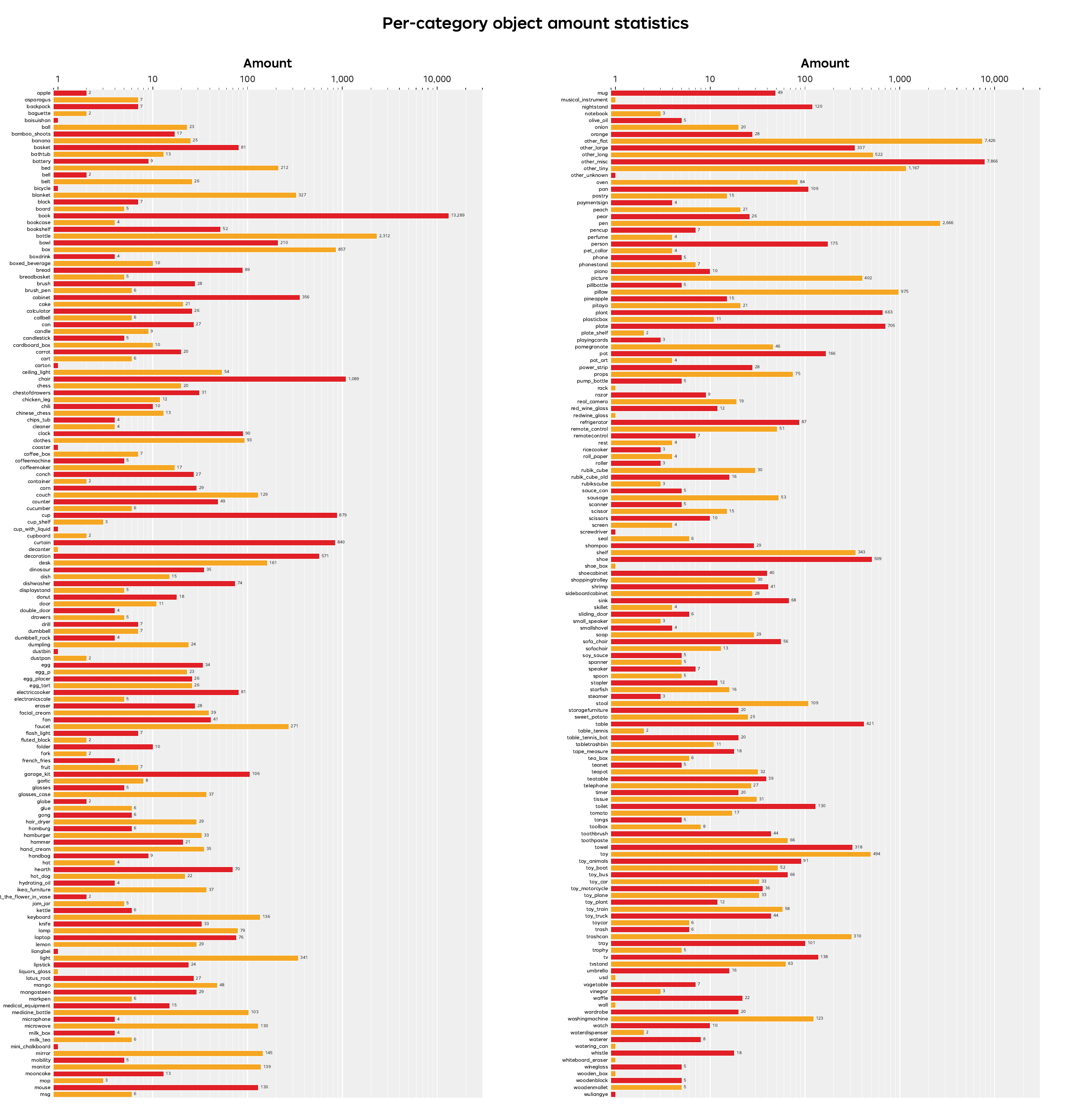}
    \caption{\textbf{Per-Category Object Amount Statistics.}
The bar chart reports the number of asset instances in each fine-grained object category of the sim-ready asset library.
It complements the hierarchical taxonomy in Figure~\ref{fig:app_sunburst_merged} by showing the instance-level scale and long-tail distribution of the collected assets.}
    \label{fig:bar_merged}
\end{figure}

This appendix summarizes the sim-ready asset library used in the simulator.
Figure~\ref{fig:app_sunburst_merged} presents its hierarchical taxonomy, while
Figure~\ref{fig:bar_merged} reports per-category instance counts.
Together, they characterize the library's semantic coverage and instance-level
diversity, supporting controlled object replacement, clutter synthesis, and
domain randomization.
Table~\ref{tab:assets} further summarizes asset categories spanning furniture,
storage objects, tools, appliances, food items, and fixtures.

\newpage

\begin{longtable}{@{}cccccc@{}}
\caption{\textbf{Sim-Ready Asset Distribution by Class and Subgroup.}}
\label{tab:assets}\\
\toprule
Top class & Count & Total share & Subgroup & Count & Class share \\
\midrule
\endfirsthead

\caption[]{Distribution of sim-ready assets by top-level class and subgroup (continued).}\\
\toprule
Top class & Count & Total share & Subgroup & Count & Class share \\
\midrule
\endhead

\midrule
\multicolumn{6}{c}{Continued on next page}\\
\endfoot

\bottomrule
\endlastfoot

Fixture        & 20{,}205 & 37.7\% & Sundry      & 17{,}320 & 85.7\% \\
               &          &        & Decoration  &  2{,}053 & 10.2\% \\
               &          &        & Sanitary    &    810   &  4.0\% \\
               &          &        & Openings    &     22   &  0.1\% \\
\midrule
Tool           & 16{,}361 & 30.5\% & Stationery  & 16{,}022 & 97.9\% \\
               &          &        & Kitchen     &    192   &  1.2\% \\
               &          &        & Hand Tool   &     98   &  0.6\% \\
               &          &        & Cleaning    &     49   &  0.3\% \\
\midrule
Storage        &  6{,}363 & 11.9\% & Containers  &  4{,}656 & 73.2\% \\
               &          &        & Boxes       &    945   & 14.9\% \\
               &          &        & Storage     &    629   &  9.9\% \\
               &          &        & Bags        &    133   &  2.1\% \\
\midrule
Textile        &  3{,}117 &  5.8\% & Bedding     &  1{,}302 & 41.8\% \\
               &          &        & Curtain     &    840   & 26.9\% \\
               &          &        & Clothing    &    653   & 20.9\% \\
               &          &        & Cloth       &    322   & 10.3\% \\
\midrule
Furniture      &  2{,}798 &  5.2\% & Seating     &  1{,}396 & 49.9\% \\
               &          &        & Tables      &    675   & 24.1\% \\
               &          &        & Shelving    &    395   & 14.1\% \\
               &          &        & Beds        &    332   & 11.9\% \\
\midrule
Appliance      &  2{,}190 &  4.1\% & Electronics &    929   & 42.4\% \\
               &          &        & Kitchen     &    676   & 30.9\% \\
               &          &        & Illumination &    481   & 22.0\% \\
               &          &        & Personal    &    104   &  4.7\% \\
\midrule
Food           &  1{,}357 &  2.5\% & Fresh       &    631   & 46.5\% \\
               &          &        & Cooked      &    364   & 26.8\% \\
               &          &        & Hygiene     &    323   & 23.8\% \\
               &          &        & Drink       &     39   &  2.9\% \\
\midrule
Entertainment   &  1{,}270 &  2.4\% & Plush       &    738   & 58.1\% \\
               &          &        & Vehicles    &    334   & 26.3\% \\
               &          &        & Game        &    124   &  9.8\% \\
               &          &        & Sport       &     74   &  5.8\% \\
\midrule
Total          & 53{,}661 & 100.0\% &             &          &        \\

\end{longtable}

\end{appendices}

\clearpage

\bibliographystyle{plainnat}
\bibliography{paper}

\end{document}